\documentclass[a4paper,12pt,fleqn]{book}

\usepackage[T1]{fontenc}
\usepackage[utf8]{inputenc}
\usepackage[french,german,english]{babel}

\usepackage{tgadventor}
\usepackage{amssymb}
\usepackage{algorithm}
\usepackage[noend]{algpseudocode}
\usepackage{algorithmicx}

\usepackage{setspace} 

\usepackage{xcolor}
\usepackage{graphicx}
\usepackage{colortbl}
\graphicspath{{images/}}
\usepackage{multirow}
\usepackage{subcaption}
\usepackage{cleveref}
\usepackage{booktabs}
\usepackage{lipsum}
\usepackage{microtype}
\usepackage{url}
\usepackage[final]{pdfpages}
\usepackage{amsmath}
\DeclareMathOperator*{\argmax}{argmax}

\usepackage{fancyhdr}

\fancypagestyle{plain}{
	\fancyhf{}

	\fancyfoot[EL,OR]{\thepage}}
\fancypagestyle{addpagenumbersforpdfimports}{
	\fancyhead{}
	
	\fancyfoot{}
	\fancyfoot[RO,LE]{\thepage}
}

\usepackage{listings}
\usepackage{hyperref}
\hypersetup{pdfborder={0 0 0},
	colorlinks=true,
	linkcolor=black,
	citecolor=black,
	urlcolor=black}
\makeatletter
\def\cleardoublepage{\clearpage\if@twoside \ifodd\c@page\else
    \hbox{}
    \thispagestyle{empty}
    \newpage
    \if@twocolumn\hbox{}\newpage\fi\fi\fi}
\makeatother \clearpage{\pagestyle{plain}\cleardoublepage}

\usepackage{color}
\usepackage{tikz}
\usepackage[explicit]{titlesec}

\titlespacing*{\chapter}{0pt}{50pt}{30pt}
\titlespacing*{\section}{0pt}{13.2pt}{*0}  
\titlespacing*{\subsection}{0pt}{13.2pt}{*0}
\titlespacing*{\subsubsection}{0pt}{13.2pt}{*0}

\usepackage{amssymb}
\usepackage{mathtools}

\begin{document}
\frontmatter
\begin{titlepage}
\begin{center}

\begin{minipage}{.4\textwidth}
    \centering
    \includegraphics[width=0.5\textwidth]{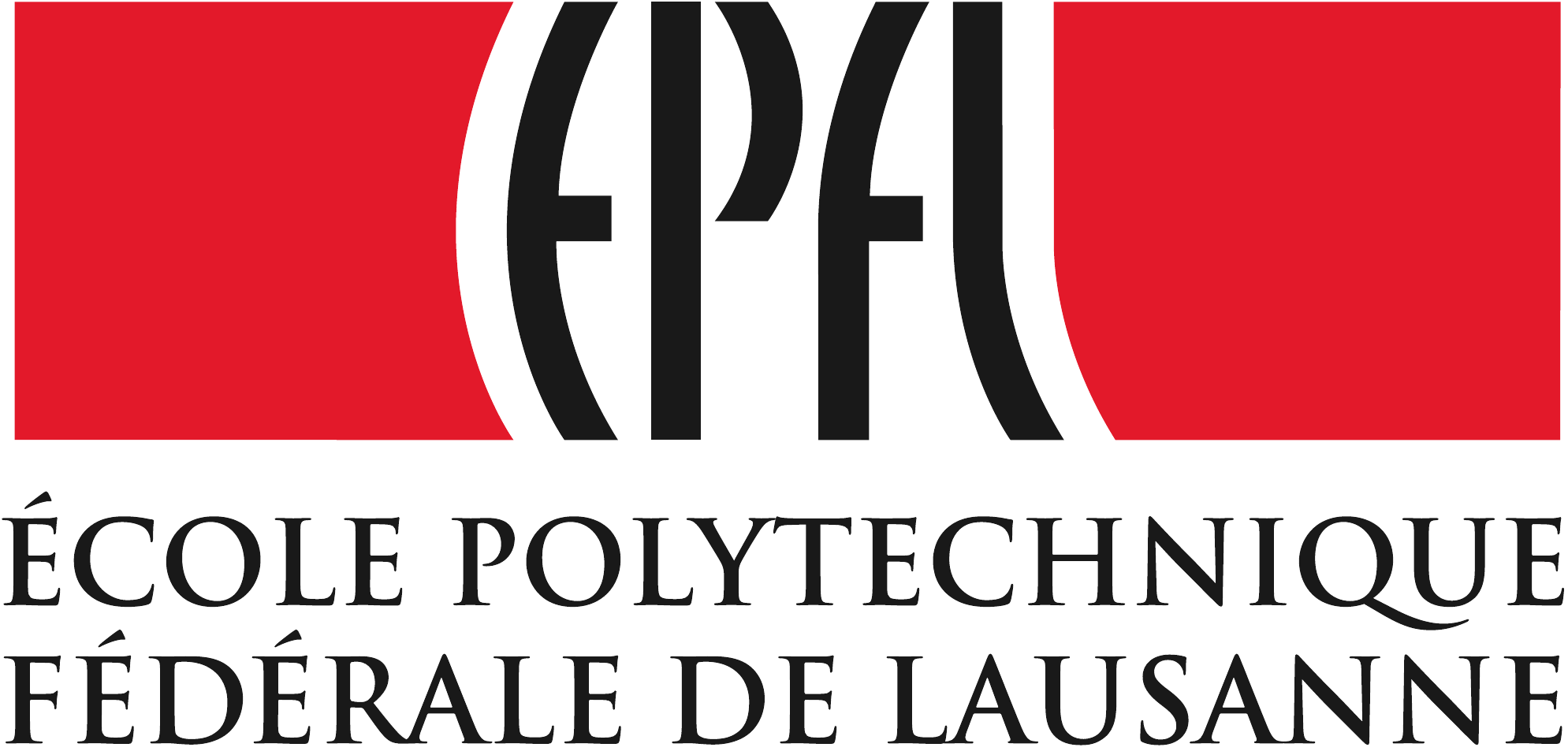}~\\[0.7cm]
    \textsc{Computer Communications and Applications Laboratory 3 (LCA3)}\\[1.5cm]
\end{minipage}
\hspace{2cm}
\begin{minipage}{0.4\textwidth}
    \centering
    \includegraphics[width=0.8\textwidth]{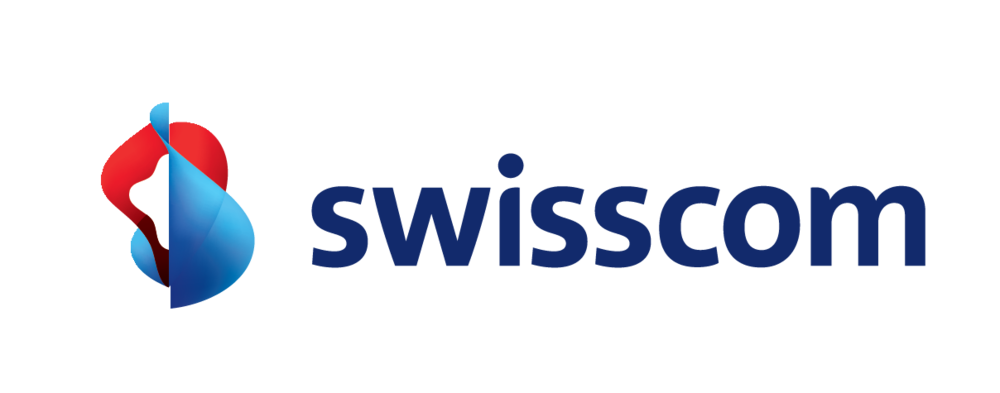}~\\[0.21cm]
    \textsc{Artificial Intelligence \& Machine Learning Group}\\[1.5cm]
\end{minipage}

\vspace{2cm}

\textsc{\Large Master Thesis}\\[1.5cm]

\noindent\rule{15cm}{0.4pt} \\ [0.4cm]
{ \huge \bfseries Building Advanced Dialogue Managers for Goal-Oriented Dialogue Systems\\[0.4cm] }
\noindent\rule{15cm}{0.4pt}

\vspace{1.5cm}

\begin{center}
    {\Large Vladimir Ilievski}
\end{center}

\vspace{2cm}

\noindent
\begin{flushleft}

\begin{minipage}[t]{0.5\textwidth}
\begin{flushleft} \large
\emph{Supervisors} \\
\vspace{0.5cm}
\begin{tabular}{ll}
EPFL:     & Prof. Patrick Thiran \\
Swisscom: & Dr. Claudiu Musat 
\end{tabular}
\end{flushleft}
\end{minipage}%
\end{flushleft}

\vfill

{\large March 16, 2018}

\end{center}
\end{titlepage}

\cleardoublepage
\chapter*{Abstract}
\addcontentsline{toc}{chapter}{Abstract} 

Goal-Oriented (GO) Dialogue Systems, colloquially known as goal oriented chatbots, help users achieve a predefined goal (e.g. book a movie ticket) within a closed domain. A first step is to understand the user's goal by using natural language understanding techniques. Once the goal is known, the bot must manage a dialogue to achieve that goal, which is conducted with respect to a learnt policy. 
The success of the dialogue system depends on the quality of the policy, which is in turn reliant on the availability of high-quality training data for the policy learning method, for instance Deep Reinforcement Learning. \\

Due to the domain specificity, the amount of available data is typically too low to allow the training of good dialogue policies. In this master thesis we introduce a transfer learning method to mitigate the effects of the low in-domain data availability. Our transfer learning based approach improves the bot's success rate by $20\%$ in relative terms for distant domains and we more than double it for close domains, compared to the model without transfer learning. Moreover, the transfer learning chatbots learn the policy up to 5 to 10 times faster. Finally, as the transfer learning approach is complementary to additional processing such as warm-starting, we show that their joint application gives the best outcomes.

\chapter*{Acknowledgements}
\markboth{Acknowledgements}{Acknowledgements}
\addcontentsline{toc}{chapter}{Acknowledgements}

First of all, I would like to express my gratitude to my supervisor at EPFL, prof. Patrick Thiran for sharing his valuable ideas and insights with me and the team over the course of this master thesis.\\

I would also like to thank my supervisor at Swisscom, Dr. Claudiu Musat for motivating, leading and supporting me during my work on this master thesis.\\

I would also like to express my sincere gratitude to my family - my parents and my sister - for supporting me throughout my studies and my life in general. I could not have imagined achieving this without their support.\\

Last but not least, I would like to thank prof. Joseph Sifakis and Dr. Simon Bliduze for giving me an opportunity to work part time as a Research Scholar at the Rigorous System Design Lab (RiSD), and support myself during the studies at EPFL.

\bigskip
 
\noindent\textit{Lausanne, 16 March 2018}
\hfill Vladimir Ilievski

\tableofcontents
\addcontentsline{toc}{chapter}{List of figures} 
\listoffigures

\setlength{\parskip}{1em}

\mainmatter

\chapter{Introduction}

Today, we live in the era of Artificial Intelligence (AI), which is penetrating in every aspect of our life. Part of this AI ecosystem are the spoken and text-based Dialogue Systems, which usage is constantly growing. These systems are quite popular because they have a potential to convey a conversation, just like a real human, acting in a truly intelligent manner. 

To increase the hype, the Loebner prize \cite{epstein1992quest} is a competition for text-based Dialogue Systems inspired by the Turing's imitation game. The aim is to stimulate and motivate the creation of truly intelligent conversational machines. Despite this, the competition has not had a winner in all previous editions. According to \cite{kurzweil2010singularity} it is only a matter of time when that will happen. 

Moreover, with the increasing pervasiveness of smart phones and ubiquitous computer systems, the Dialogue Systems are getting even more attractive. Consequently, they started being used in a plethora of different applications, ranging from trivial chit-chatting to personal assistants. An example of such systems are the popular Apple's Siri, Google Now and Cortana from Microsoft \cite{strayer2017smartphone}. Therefore, it is of paramount importance to continue the development of these systems and push the boundaries even further.

The text-based Dialogue Systems colloquially known as Chatbots,
are divided in two groups, depending on the nature of the conversation. In fact, there are: $i)$ open-domain and $ii)$ closed-domain Chatbots.


In the open-domain setting the conversation can go in any direction, which means the users can have an open conversation with the Chatbot about everything, usually in the format of chit-chatting, without any or minimal functionality. For instance, \cite{serban2016building} used the Movie-DiC \cite{banchs2012movie} corpus of movie dialogues to build a general-purpose Chatbot. Because of the general-coverage nature, the open-domain Chatbots require huge amount of annotated data, thus making it almost impossible to create one.

On the other hand, the closed-domain Dialogue Systems are more practical and easier to implement, because they focus only on few aspects and are designed to help users to achieve predetermined goals in a predefined domains. For example, it could be a travel planning task \cite{peng2017composite} or restaurant table booking dialogue system \cite{wen2016network}, to help users book a flight or table in restaurant, in the most convenient way, i.e. by conversation. For this reason they are called Goal-Oriented (GO) Chatbots and can be grouped together in larger systems such as Amazon Alexa\footnote{https://developer.amazon.com/alexa} to give an impression of a general coverage. Each individual component (which in Amazon Alexa can be viewed as skills of the overarching generalist bot) is closed-domain in nature.


\section{Contributions and Thesis Outline}

This thesis focuses only on a subset of the Goal-Oriented Chatbots, modeled as Partially Observable Markov Decision Processes (POMDP) \cite{young2013pomdp}, where the rich set of Reinforcement Learning (RL) algorithms \cite{sutton1998reinforcement} can be used  to train them, for instance the Deep Q-Nets (DQN) \cite{mnih2015human}. The lack of in-domain dialogue data is a key problem for training high quality RL-based Goal-Oriented Chatbots. We need in-domain labeled dialogues for two reasons: $i)$ to warm-start the Chatbot, which is a standard widely used technique and $ii)$ to train the chatbot by simulating a considerable number of different conversations.

In this thesis we argue that the domain similarity can be leveraged in a clever way to build efficient GO Chatbots with less data, using the so-called \textit{Transfer Learning} technique. We use the similarity between a \textit{source} and a \textit{target} domain, as many domains, such as restaurant and movie booking, share to a large extent some common information. For example, in the restaurant booking scenario, the user might ask the question \textit{``Which restaurant I can book a table for 3 people for tomorrow?''}, while in the movie booking domain the question could be \textit{``Which theater I can book 3 tickets for tomorrow?''}. In both domains, the user includes information for number of people and time. We believe this information need not be learnt twice and that a transfer is possible. 

We successfully combine \textit{Transfer Learning} and RL-based Goal-Oriented Chatbots and to the best of our knowledge we are the first ones doing that. As a result of the research over the course of this thesis, we published a paper \cite{ilievski2018goal}, which is submitted for a review at the 27th International Joint Conference on Artificial Intelligence (IJCAI). The contributions of this thesis are following:
\begin{itemize}
	\item\textbf{Training GO Chatbots with less data}: In data constrained environments, models trained with \textit{Transfer Learning} achieve better training and testing performances than ones trained independently.
    
    \item\textbf{Better GO Chatbot performance}: Using \textit{Transfer Learning} has a significant positive effect on performance even when all the data from the target domain is available.
    
    \item\textbf{Intuitions on further improvements}: We show the gains obtained with \textit{Transfer Learning} are complementary to the ones due to warm-starting and the two can be successfully combined.
    
    \item\textbf{New published datasets}: We publish new datasets for training Goal-Oriented Dialogue Systems, for restaurant booking and tourist info domains\footnote{The datasets will be published in the camera-ready version}. They are derived from the third Dialogue State Tracking Challenge~\cite{henderson2013dialog}.

\end{itemize}

After the Related Work Chapter (Chapter \ref{chap:related_work}), we first make a general overview of the Dialogue Systems in Chapter \ref{chap:dial_sys}. Then, in Chapter \ref{chap:model_rl_based}, we describe the model of the RL-based GO Chatbots, which performance relies on a robust dialogue state tracking and an efficient learnt policy, as described in Chapter \ref{chap:efficient_policy_learning}. We conduct our experiments and show the results in Chapter \ref{chap:experiments_results}. Finally in Chapter \ref{chap:conclusion} we conclude our work and present the possible future work.

\chapter{Related Work}
\label{chap:related_work}

\section{Goal-Oriented (GO) Dialogue Systems}

The Goal-Oriented (GO) Dialogue Systems have been under development in the past two decades, starting from the basic, handcrafted Dialogue Systems. For instance, \cite{larsson2000information} introduced a framework for dialogue management development, based on hand-crafted rules. In the same direction, \cite{zue2000juplter} built a weather information, hand-crafted Goal-Oriented Chatbot. The recent efforts to build such systems are generally divided in three lines of research.

\subsection{Fully-Supervised Models}

The first way is to treat them in an end-to-end, fully supervised sequence-to-sequence manner \cite{sutskever2014sequence}. Thus, we can use the power of the deep neural networks based on the encoder-decoder principle to infer the latent representation of the dialogue state. However, it is worth noting that these models require considerable amount of data.

The authors in \cite{vinyals2015neural} used standard Recurrent Neural Networks (RNNs) and trained a Goal-Oriented Chatbot in a straightforward sequence-to sequence fashion. They benchmarked their findings in IT helpdesk troubleshooting domain, where costumers face computer related issues, and a specialist help them by conversing and walking through a solution. Due to the incapability of the recurrent nets to compress very long dependencies, this chatbot is not having a strong reasoning power. 

To overcome the RNN memory limitations, in their work \cite{bordes2016learning} used LSTM cells in combination with explicit memory, known as Memory Networks \cite{sukhbaatar2015end}, to build a Goal-Oriented Chatbot, using the bAbI\footnote{https://research.fb.com/downloads/babi/} tasks for restaurant reservation. The chatbot demonstrated better reasoning power, thus remembering and updating the past user preferences. For this reason, this work represents a testbed for testing the shortcomings and strengths of fully-supervised end-to-end Goal-Oriented Dialogue Systems.


\subsection{Reinforcement Learning-based Models}

Another branch of research had emerged, focusing on the Deep Reinforcement Learning because the fully-supervised approach is data-intensive. These models have a quite complex structure, since they include many submodules, such as Natural Language Understanding (NLU)~\cite{hakkani2016multi} and Natural Language Generation (NLG)~\cite{wen2015semantically} units, as well as a Dialogue State Tracker (DST).

Aligned in this direction, \cite{cuayahuitl2017simpleds} created a simple Dialogue System for a restaurant reservation domain. The system's actions solely depend on the RL-based agent, by performing action selection directly from raw text of the last user and system responses instead of manual feature engineering. Therefore, this system does not include any language understanding and state tracking units, which in turn is quite constraining. Another simple Question-Answering Chatbot (Q\&A bot) is presented in \cite{dhingra2016end}. It is an end-to-end 
Reinforcement Learning Chatbot, which helps users search Knowledge Bases (KBs) for movies, without composing complicated queries.

As an extension of all the previous work, \cite{li2017end} went one step further and built a comprehensive Movie Booking Chatbot. It includes a User Simulator \cite{li2016user}, which simulates the user in the training process, Natural Language Understanding (NLU) and Natural Language Generation (NLG) units, as well as a basic Dialogue State Tracker and a Policy Learning module. It is trained in an end-to-end fashion, by leveraging the Deep Q-Nets (DQN) \cite{mnih2015human} for the policy learning.

\subsection{Hybrid Models}

This line of research combines both, the Fully-Supervised and Reinforcement Learning-based approaches, in order to escape the limitations, characteristic for both of them.

In their work ~\cite{su2016continuously} described a two-step approach to train a policy for Goal-Oriented Chatbot. In the first step, the algorithm is trained on a fixed corpus data in a supervised way. In the second step, the policy is fine-tuned using RL-based policy gradient \cite{williams1992simple} technique, in order to explore the dialogue space more efficiently.

Similarly, \cite{williams2017hybrid}, proposed a solution to
train a Goal-Oriented Dialogue System in two modes: off-line and on-line mode. In the off-line mode, the system is trained in a fully-supervise manner, combining an LSTM network with hand-crafted templates, to mitigate the data requirements. Afterwards, in the on-line mode, the system learns autonomously by incorporating RL-based policy gradient approach. 

\section{Data-Constrained Dialogue Systems}

One desired property of the Goal-Oriented Chatbots is the ability to switch to new domains and at the same time not to lose any knowledge learned from training on the previous ones. This property is enforced due to the lack of in-domain data required to train high-quality Goal-Oriented Chatbots.

In this direction, the authors in \cite{gavsic2015distributed} proposed a Gaussian Process-based technique to learn generic dialogue polices, which are organized in a class hierarchy. These policies with a modest amount of data can be furthermore adjusted according to the use case of the dialogue system.


On the other hand, \cite{wang2015learning} learned domain-independent dialogue policies, such that they parametrized the ontology of the domains. In this way, they show that the policy optimized for a restaurant search domain can be successfully deployed to a lap-top sale domain.


Last but not the least, ~\cite{lee2017toward} utilized a continual learning technique, to smoothly add new knowledge in the neural networks that specialized a dialogue policy in an end-to-end fully-supervised manner.  


Nevertheless, none of the previously mentioned papers tackles the problem of transferring the domain knowledge in case when the dialogue policy is optimized using a Deep Reinforcement Learning. In this thesis, we propose such method, based on the standard Transfer Learning technique~\cite{pan2010survey}. Therefore, using this method we surpass the limitations to transfer the in-domain knowledge in Goal-Oriented Dialogue Systems based on Deep Reinforcement Learning.

\chapter{Overview of Dialogue Systems}
\label{chap:dial_sys}

This chapter gives an overview of the Dialogue Systems in general. The research on the dialogue systems intents to create comprehensive systems that can hold a real conversation, successfully employing all aspects: reasoning power, giving well defined and reasonable responses, emotion detection etc. Depending on the nature of the input and output, there are two types of Dialogue Systems: $i)$ Spoken Dialogue Systems (SDS) and $ii)$ Text-Based Dialogue Systems colloquially known as Chatbots. Fundamentally, both types exhibit many similarities, only the pre- and post-processing techniques differ.

There is no consensus on the architecture of the Dialogue Systems, it is case-dependent. In general they are always composed of two parts: $i)$ user, whether real or simulated and $ii)$ the internal system. One good reference is given in \cite{pieraccini2005we}. In any case, both parts converse in alternating manner such that a \textit{dialogue turn} is one cycle of consecutive utterances from the user and the system.

\section{Spoken Dialogue Systems}

\begin{figure}[tb] 
\centering 
\includegraphics[width=\columnwidth]{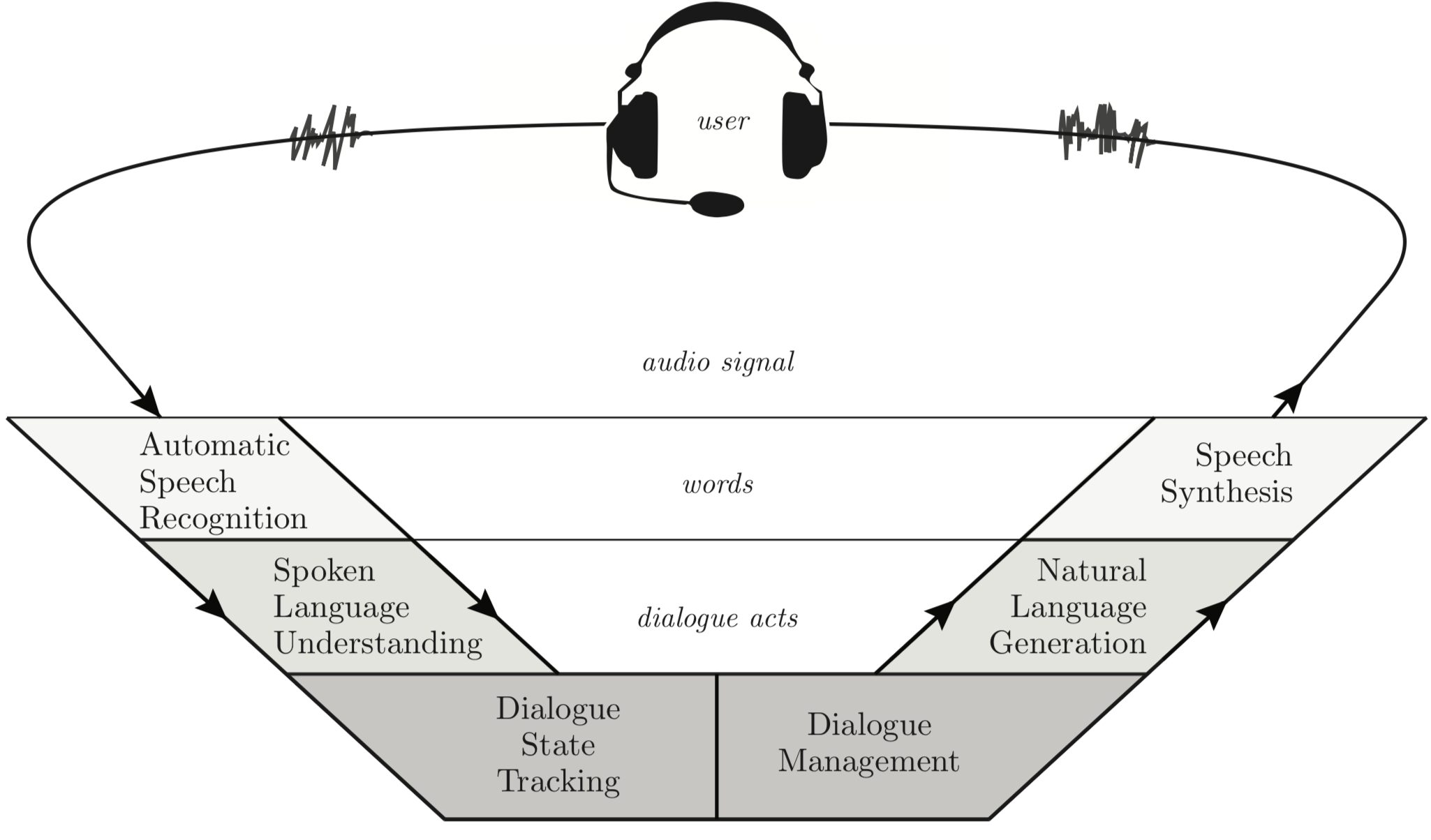} 
\caption{The architecture of a Spoken Dialogue System from \cite{henderson2015discriminative}}
\label{fig:sds_pipeline} 
\end{figure}

The \textit{Spoken Dialogue Systems} are specially designed for environments which does not include user interfaces such as big screens and keyboards, they rather use microphone and speakers. \cite{henderson2015discriminative} shows a typical composition of the Spoken Dialogue Systems as well as the information flow. For this reason, the input and output is continuous speech signal, which requires special modules and techniques to handle  (see Figure \ref{fig:sds_pipeline}), which includes:
\begin{itemize}
\item The \textit{Automatic Speech Recognition} (ASR) unit \cite{zhang2017towards} assigns probabilities to the words in the user utterance.
\item The \textit{Spoken Language Understanding} (SLU) unit \cite{yao2014spoken} infers the semantics of the user input.
\item The \textit{Speech Synthesis} (SS) unit \cite{zen2009statistical}, will convert the system's response into speech.
\end{itemize}






\section{Text-Based Dialogue Systems (Chatbots)}

Contrary to the Spoken Dialogue Systems, the Text-Based Systems (known as Chatbots) only focus on the interfaces which include screens and keyboards, meaning the input and output is text. Consequently, they include appropriate units and algorithms to handle text. Depending on the nature of the conversation, the Chatbots are divided in Open-Domain and Closed-Domain Chatbots known as Goal-Oriented (GO) Chatbots.

In the open-domain setting, the conversation can go in any direction, usually in the form of chit-chatting, without any purpose. Because of their nature to cover every possible case, it is almost impossible to create a perfect open-domain Chatbot, as shown in \cite{serban2016building}. Thus, most of the chatbot research is on the closed-domain Chatbots, which is a case in this thesis.

\section{Goal-Oriented (GO) Chatbots}

The Goal-Oriented (GO) Dialogue Systems are more useful and practical and are easier to implement. This is due to the fact that their domain of expertise is much narrower, focusing only on few key points of the dialogue. In general there are two dominant paradigms in GO Dialogue Systems implementations: $i)$ Fully-Supervised and $ii)$ Reinforcement Learning (RL) based.

\subsection{Fully-Supervised GO Chatbots}
In case of fully supervised implementation, we apply the recurrent neural networks (RNNs) encoder-decoder principles, mainly applied in the machine translation. An example of this kind of models is presented in \cite{bordes2016learning,wen2016network}. These models are trained in a sequence-to-sequence fashion \cite{sutskever2014sequence}, that encode the user request and its context and decode the bot answer directly.

The fully supervised GO chatbots require a considerable amount of annotated human-human or human-machine dialogues since the system is trying to mimic the knowledge of the expert. Moreover, we don't have a control over the internal state, which means we can not model the dialogue as we wish.

\subsection{Reinforcement Learning (RL) based GO Chatbots}
On the other hand, we can model the GO Chatbots as a Partially Observable Markov Decision Process (POMDP)\cite{young2013pomdp}. The Reinforcement Learning (RL) \cite{sutton1998reinforcement} algorithms are one rich subset of powerful and promising algorithms that can be applied in this case.

\begin{figure}[t!]
    \centering
    \includegraphics[width=\textwidth]											{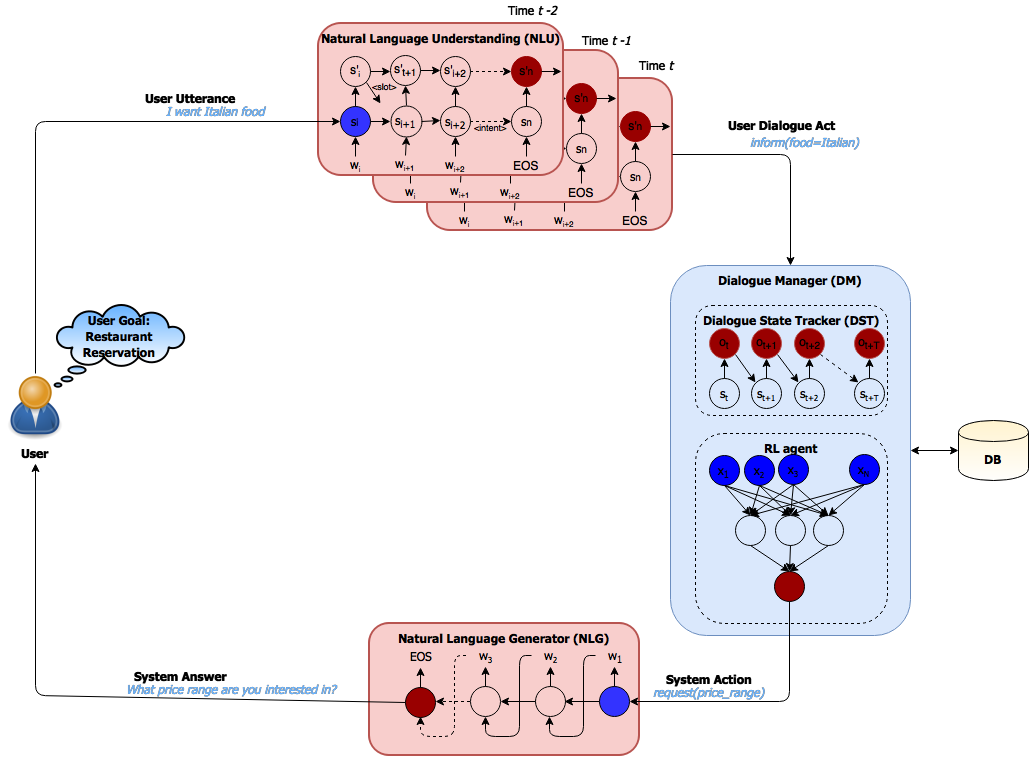}
    \caption{Text-Based Dialogue System modeled as a Partially Observable Markov Decision Process. The user utterance is parsed by the NLU unit producing a dialogue act understandable for the system. In the Dialogue Manager, the state tracker is estimating the state such that the RL agent could take the ideal action. This action is further passed to the NLG unit and finally presented to the user in a human readable form.}
    \label{fig:dialogue_systems}
\end{figure}

Figure \ref{fig:dialogue_systems} shows a typical composition of an RL-based GO Chatbot, as well as the information flow. Following the pipeline, there are three separate components, each having a specific role in the process:

\begin{enumerate}
\item First of all the \textit{Natural Language Understanding} unit \cite{hakkani2016multi} will infer the semantics of the user input. This includes understanding the user intent and the slots (i.e. the relevant information).

\item In the second step, the \textit{Dialogue Manager} (DM) will take care of the dialogue. Based on the context, previous user and system actions it will produce the next system action. Usually it includes two other subcomponents: $i)$ \textit{Dialogue State Tracker} (DST) \cite{henderson2015machine}, which purpose is to build reliable state of the dialogue and a \textit{Policy Learning} module which reads the dialogue states and takes the next system action.

\item Finally, the \textit{Natural Language Generation} unit \cite{wen2015semantically}, based on the DM output, will generate a natural sentence, understandable for the end user.
\end{enumerate}

RL-based Chatbots require less annotated dialogues than their sequence-to-sequence counterparts, due to their ability to simulate the conversation, thus exploring the unknown dialogue space more efficiently. The data requirements are however not trivial and obtaining the dialogue data is still the biggest obstacle their creators face. In the following chapters we dive deeper in the RL-based GO Chatbots and show how this obstacles can be surpassed using \textit{Transfer Learning}.

\chapter{Model of RL-based GO Chatbots}
\label{chap:model_rl_based}

If we model the GO Dialogue Systems as POMDP and apply the Reinforcement Learning techniques as in \cite{zhao2016towards,li2017end}, then, several components are compromising that system as shown in Figure  \ref{fig:dialogue_systems}. It consists of two independent units: the \textit{User Simulator} on the left side and the \textit{Dialogue Manager} (DM) on the right side. In-between there are the Natural Language Understanding (NLU) and Natural Language Generator (NLG) units. Our work is based on the model from~\cite{li2017end}, who proposed an end-to-end reinforcement learning approach to build a Goal-Oriented Chatbot in a movie booking domain.

Goal-Oriented bots contain an initial NLU component, that is tasked with determining the user's \textit{intent} (e.g. \textit{book a movie ticket}) and its parameters, also known as \textit{slots} (e.g. date: \textit{today}, count: \textit{three people}, time: \textit{7 pm}).
The usual practice in the RL-based Goal-Oriented Chatbots is to define the user-bot interactions as \textit{semantic frames}. At some point $\mathbf{t}$ in the time, given the user utterance $\mathbf{u_{t}}$, the system needs to perform an action $\mathbf{a_{t}}$. A bot action is, for instance, to request a value for an empty slot or to give the final result.

The entire dialogue can be reduced to a set of \textit{slot-value} pairs, called \textit{semantic frames}. 
Consequently, the conversation can be executed on two distinct levels:

\begin{enumerate}
	\item \textbf{Semantic level:} the user sends and receives only semantic frames as messages.
    \item \textbf{Natural language level:} the user sends and receives natural language sentences, which are reduced to, or derived from a semantic frame by using Natural Language Understanding (NLU) and Natural Language Generation (NLG) units respectively~\cite{wen2015semantically,hakkani2016multi}.
\end{enumerate}

For instance, in the movie booking domain one semantic frame could be defined as:
\makebox[\textwidth]{\textit{\{movie\_name:  ``Titanic'', number\_of\_people:  ``2'', theater, intent=``request''\}, }}

which in natural language could be written as:\\

\makebox[\textwidth]{\textit{``Which theater I can book 2 tickets for the movie Titanic?''}}

By exchanging this kind of a data structures, the user and the system can convey the entire conversation until reaching the goal.

\section{User Simulator}

The \textit{User Simulator} creates a user - bot conversation, given the semantic frames. Because the model is based on Reinforcement Learning, a dialogue simulation is necessary to successfully train the model \cite{li2016user}.

From the dataset of available user goals the User Simulator randomly picks one, which is unknown for the Dialogue Manager. The user goal consists of two different sets of slots: \textit{inform slots} and \textit{request slots}. 
\begin{itemize} 
\item \textit{Inform slots} are the slots for which the user knows the value, i.e. they represent the user constraints (e.g. \{movie\_name:  ``avengers'', number\_of\_people:  ``3'', date:  ``tomorrow''\}). 
\item \textit{Request slots} are ones for which the user is looking for an answer (e.g. \{ city, theater, start\_time \} ). 
\end{itemize}
Having the user goal as an anchor, the user simulator generates the \textit{user utterances} $\mathbf{u_{t}}$. The initial user utterance, similar to the user goal, consists of the initial inform and request sets of slots.
Additionally, it includes a user intent, like \textit{open dialogue} or \textit{request additional info}.

The user utterances generated over the course of the conversation follow an agenda-based model~\cite{schatzmann2009hidden}. According to this model, the user is having an internal state $\mathbf{s_{u}}$, which consists a goal $G$ and an agenda $A$. The goal furthermore is split in user constraints $C$ and user requests $R$. In every consecutive time step $\mathbf{t}$, the user simulator creates the user utterance $\mathbf{u_{t}}$, using its current state $\mathbf{s_{u}}$ and the last system action $\mathbf{a_{t}}$. In the end, using the newly generated user utterance $\mathbf{u_{t}}$, it updates the internal state $\mathbf{s^{\prime}_{u}}$.

\section{Natural Language Understanding Unit}

The \textit{NLU} unit is responsible for transforming the user utterance to a predefined \textit{semantic frame} according to the system's conventions, i.e. to a format understandable for the system. This includes a task of slot filling and intent detection.

For example, the intent, could be a \textit{greeting}, like \textit{Hello}, \textit{Hi}, \textit{Hey}, or it could have an \textit{inform} nature, for example \textit{I like Indian food}, where the user is giving some additional information. Depending on the interests, the slots could be very diverse, like the \textit{actor name}, \textit{price}, \textit{start time}, \textit{destination city} etc. As we can see, the intents and the slots are defining the closed-domain nature of the Chatbot.

The task of slot filling and intent detection is seen as a sequence tagging problem. For this reason, the NLU component is usually implemented as an LSTM-based recurrent neural network with a Conditional Random Field (CRF) layer on top of it. The model presented in \cite{hakkani2016multi}, is a sequence-to-sequence model using bidirectional LSTM net, which fills the slots and predicts the intent in the same time. On the other hand, the model in \cite{liu2016attention} is doing the same using an attention-based RNN. 

To achieve such a task, the dataset labels consist of: concatenated B--I--O (Begin, Inside, Outside) slot tags, the intent tag and an additional end-of-string (EOS) tag. As an example, in a restaurant reservation scenario, given the sentence \textit{Are there any French restaurants in Toronto downtown?}, the task is to correctly output, or fill, the following slots: \textit{\{cuisine: French\}} and \textit{\{location: Toronto downtown\}}. The table below shows how we would correctly tag the previous example. One very effective technique to build better NLU units with less data (based on the \textit{Active Learning} methodologies) is presented in \cite{dimovski2018submodularity}.

\begin{table}[h!]
\centering
\caption{An example of tagging a sentence in B--I--O (Begin, Inside, Outside) format}
\label{table:bio_example}
\begin{tabular}{|l|l|l|l|l|l|l|l|}
\hline
\rowcolor[HTML]{C0C0C0} 
Are & there & any & French    & restaurants & in & Toronto    & downtown?  \\ \hline
O   & O     & O   & B-Cuisine & O           & O  & B-Location & I-Location \\ \hline
\end{tabular}
\end{table}

\section{Natural Language Generator Unit}
The NLG unit, on the other hand is the glue between the system and the user. Given the system response as a \textit{semantic frame}, it maps back to a natural language sentence, understandable for the end user. The \textit{NLG} component can be rule-based or model-based. In some scenarios it can be a hybrid model, i.e. combination of both.

The rule-based NLG outputs some predefined template sentences for a given \textit{semantic frame}, thus they are very limited without any generalization power. For this reason, they are only used in special occasions.

On the other hand, the model-based NLG units, are having learnable parameters and are usually trained in a sequence-to-sequence fashion. The models presented in \cite{wen2016conditional,wen2015semantically}, use an LSTM-decoder with a given \textit{semantic frame}, to generate template-like sentences with slot placeholders. Afterwards, a beam searched is applied, to replace the placeholders with actual values.

\section{Dialogue Manager}

At the core of the GO Dialogue Systems lies the \textit{Dialogue Manager} (DM), supported by the NLU and NLG units. Additionally, the DM could be connected to some external Knowledge Base (KB) or Data Base (DB), such that it can produce more meaningful answers.

The Dialogue Manager consists the following two components: the \textit{Dialogue State Tracker} (DST) and the \textit{Policy Learning} which is the RL agent.

The \textit{Dialogue State Tracker} (DST) is a complex and essential component that should correctly infer the belief about the state of the dialogue, given all the history up to that turn. The \textit{Policy Learning} is responsible for selecting the best action, i.e. the system response to the user utterance, that should lead the user towards achieving the goal in a minimal number of dialogue turns.

\subsection{Dialogue State Tracker}

The \textit{Dialogue State Tracker} (DST) is producing a meaningful state $s_{t}$ of the dialogue up to time $t$ in the time. The state $s_{t}$ is a data structure, that should depict the state of the conversation to a level of detail that provides all necessary information in order an intelligent agent to easily and reliably select the next action.

The tracker, takes all possibly observable input up to time $t$ which includes: all the user utterances and system actions taken so far, all the results from the \textit{NLU} unit. Additionally, it might include all external knowledge provided in a knowledge base or a data base. For example in a restaurant search scenario, the state might indicate the user price range and cuisine preferences, what information they are seeking, like a telephone number or address.

Therefore, given all of this information, a robust dialogue state tracker outputs a distribution $p\left( s \right)$ over all possible dialogue states. This is due to the fact that the true state is not fully observable from the raw input. Several factors contribute for that: ambiguous or not clearly specified user utterances, the noise and the error from the \textit{NLU}, changes in user goal etc.

In order to tackle these challenges, in the literature there are three types of state trackers: \textit{rule-based}, \textit{generative models} and \textit{discriminative models}. Very recently, the series of the Dialogue State Tracking Challenge (DSTC) have started \cite{the-dialog-state-tracking-challenge-series-a-review,henderson:ml-for-dst-review}, a competition aiming to boost the state trackers to the next level. Many state-of-the-art dialogue state trackers emerged from this competitions.

\subsubsection{Hand-Crafted DST models}

The \textit{hand-crafted} dialogue state trackers are the most basic ones and they were used in the early dialogue systems. They infer the state of the dialogue by a manually designed and tuned parameters, such that the new state $s^{\prime}$ is derived from the last state $s$ using the last user utterance. An example of such system is the weather information system developed at MIT, called JUPITER \cite{zue2000juplter}. Moreover, in \cite{larsson2000information} a hand-crafted rules are used to build a complex dialogue management system.

One strong advantage of this kind of state trackers is that they do not require any training data. However, due to the lack of flexibility and inability to adapt and generalize over many possible states, a data-driven approach is required and obvious.

\subsubsection{Generative DST models}

\begin{figure}[h]
    \centering
    \includegraphics[width=0.3\textwidth]											{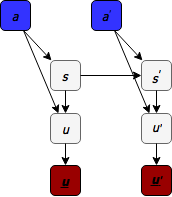}
    \caption{Dynamic Bayesian Net for inferring the state of the dialogue. The state $s^{\prime}$ depends on the previous state $s$ and the system action $a^{\prime}$. This leads to a new user utterance $u^{\prime}$ (depending both on the state and system action), which becomes noisy afterwards.}
    \label{fig:dbn_dst}
\end{figure}

For this reason, the \textit{generative models} emerged, modeling the dialogue as a dynamic Bayesian network considering the state $s$, and the user action $u$ as an unobservable random variables. In general, given the input vector $x \in \mathbb{R}^{N}$ for some $N \in \mathbb{N}$, and the label $y$, the generative models are trying to estimate the joint distribution of $x$ and $y$, i.e. $\Pr \left(y, x \right)$. Thus, the rest of the probabilities can be derived using the Bayesian rules.
The Bayesian net for inferring the new state $s^{\prime}$ is shown in Figure \ref{fig:dbn_dst}. The new state $s^{\prime}$, depends on the previous state $s$, and the current machine action $a^{\prime}$, which on the other side depends on the noisy user action $\textbf{\underline{u}}$.

\subsubsection{Discriminative DST models}

Finally, the \textit{discriminative models} for state tracking are the most powerful ones. They are directly estimating the conditional probability $\Pr \left(y | x \right)$, where $y$ is the label and $x$ is the underlying data. One example and very successful discriminative model is the model described in \cite{henderson2014word,Henderson2014d}. Using a word-based approach, the authors successfully scaled the model to work on unseen slots and values. This is done by using the \textit{n-gram} technique on top of the slot-value pairs. Moreover, in order to make the system slot invariant, \textit{delexicalized} features are used, which means introducing generic symbols. Afterwards, a Recurrent Neural Network is used to discriminate between the dialogue states. 


\subsection{Policy Learning}
The \textit{Policy Learning} module selects the next system actions to drive the user towards the goal in the smallest number of steps. It does that by using the Reinforcement Learning theory.

The theory of Reinforcement Learning is motivated by the neuroscientific perspectives of the animal and human behavior, deeply and well rooted in the nature. Therefore, by mathematically modeling the nature, we have an intelligent agent acting in an environment and perceiving a state $s$. Upon taking an action $a$, based on the policy learned from the past experiences, it is receiving a reward $r$ and it changes the state of the environment.

Therefore, the role of the dialogue agent is to learn an optimal policy for conducting efficient and successful dialogue with the user. This is done by following the reinforcement way of learning, defining final, and immediate reward, such that the agent should maximize the cumulative future reward. One way to do so, is by applying on off-policy method, such as Q-learning \cite{watkins1992q}. The Q-function is the utility of taking an action $a$, when the agent is perceiving the state $s$, by following a policy $\pi = P(a | s)$. The utility measure is defined as the problem of maximizing the cumulative future reward that the agent will receive.

The optimal action $a$, in a given state $s$, in a given time point $t$, according to the Q-learning is defined as:

\begin{equation}\label{eq:Q_func}
Q^{*} \left( s, a\right) = \max_{\pi}\mathbb{E}\left[ r_{t} + \gamma r_{t+1} + \gamma^{2} r_{t + 2} + \cdots | s_{t} = s, a_{t} = a, \pi \right],
\end{equation}

where $r_{t}, r_{t+1}, \ldots $ are the rewards at each time step, $\gamma \in [0,1]$ is the discount factor, i.e. the relevance of the future rewards, and $\pi = P(a|s)$ is the agent's policy. The optimal action-value (or Q) function obeys an important identity known as the Bellman equation, which states:

\begin{equation}\label{eq:Bellman_Eq}
Q^{*} \left( s, a\right) = r + \gamma \max_{a^{\prime}} Q^{*} \left( s^{\prime}, a^{\prime}\right).
\end{equation}

Obviously, this agent would follow a greedy strategy, and will always exploit the same set of actions in order to reach the goal. In practice, we want the agent to generalize well over the state space. For this reason, the agents incorporate different exploitation-exploration strategies. The most popular and quite effective one is the $\epsilon$ - greedy strategy, for $\epsilon \in [0, 1]$. That means with a probability of $\epsilon$ the agent will select a random action, while with a probability of $1 - \epsilon$ it will follow a greedy approach.

However, by following the Bellman Equation \ref{eq:Bellman_Eq}, the objective function for learning the Q-function would be:

\begin{equation} \label{eq:objective_func}
\mathbb{E}_{s,a,r,s^{\prime}}\left[ \left( \overbrace{ r + \gamma \max_{a^{\prime}} Q\left( s^{\prime}, a^{\prime}\right)}^\text{target value} - \overbrace{ Q\left( s, a\right)}^\text{old value} \right)^{2}  \right].
\end{equation}

Therefore, in order to find a function approximation of the Q-function, we have to minimize the equation \ref{eq:objective_func}.

The recently developed method, by a group of researches at Deep Mind, successfully applies a deep feed-forward neural network as a function approximator of the Q-function. Detailed information about the Deep Q-learning (DQN) technique and how to build more efficient policies using them will be provided in Chapter \ref{chap:efficient_policy_learning}.

\chapter{Efficient Policy Learning via Transfer Learning}
\label{chap:efficient_policy_learning}

In Chapter \ref{chap:model_rl_based}, we introduced the Q-learning and the very recent, popular and effective technique for approximating the Q-function using a deep feed forward neural network, know as Deep Q-Learning (DQN). In this chapter, we will dive in depth, explaining the advantages of the DQN, since we use it extensively in our dialogue systems.

\section{Deep Q-Learning (DQN)}

There are two types of DQN algorithms: $i)$ standard DQN and $ii)$ Double DQN (DDQN) which is an extension and more robust version of the standard DQN algorithm.

\subsection{Standard DQN}
The agents should be able to generalize well, over a high-dimensional, partially observable and complex input. Exactly this was the difficulty that most of the RL algorithms were facing, so their applicability was mainly in domains of fully observable, finite and low-dimensional states. However, all of this changed after introducing the Deep Q-Network, by a group of researchers at Deep Mind \cite{mnih2015human}. It is a standard deep feed-forward neural network, which approximately calculates $Q \left( s, a \right | \theta)$, where $\theta$ are the parameters, (i.e. weights) of the Q-Network. 

As we already explained, the goal in the reinforcement learning is to minimize the equation \ref{eq:objective_func}. However, in the RL community it is widely known that a nonlinear approximator of the Q-function, such as a neural network, causes instability and divergence. This is due to the following two reasons:

\begin{enumerate}
\item The correlation between the sequence of observations, i.e. every next state depends on the previous states and actions, and
\item The targets (labels), depend on the network weights. More precisely, to calculate the target $r + \gamma \max_{a^{\prime}} Q \left( s^{\prime}, a^{\prime}\right)$, used as a correction, we used the same weights which are changing through the time. This is in total contrast with the supervised learning, where the targets are fixed before the learning starts.
\end{enumerate}

The first problem is solved by using a biologically inspired mechanism, called \textit{experience replay}, i.e. to learn from experiences from an arbitrary point in the past. In order to perform such a mechanism, we store the agent's experiences $e_{t} = \left( s_{t}, a_{t}, r_{t}, s_{t+1} \right)$ in a dataset $D_{t} = \left\{e_{1}, \cdots , e_{t}\right\}$ of tuples. Therefore, at training time, we randomly sample experiences from the dataset $D_{t}$, following some probability distribution, which in the simplest case is a uniform distribution. Using this technique, we overcome the first problem, but we are still not able to perform learning in the network, due to the second issue, the targets still depend on the network weights.

For this reason, we introduce another Q-Network, called \textit{target network}, with fixed parameters $\theta^{\prime}$. As its name suggests, this network is only used to calculate the targets $Q^{\star} \left( s^{\prime}, a^{\prime} | \theta^{\prime} \right)$, independently from the primary Q-Network. The target network parameters are only updated with the primary Q-Network parameters every $C$ steps and are held fixed between individual updates.

Thus, with the randomly drawn mini-batch of experiences and the target Q-Network we perform learning using the following loss function:

\begin{equation}
\label{eq:loss_function}
\mathcal{L}(\theta) = \mathbb{E}_{s,a,r,s^{\prime}}\left[ \left( r + \gamma \max_{a^{\prime}} \overbrace{Q\left( s^{\prime}, a^{\prime} | \theta^{\prime} \right)}^{\text{calc. by target net}} -  \overbrace{Q\left( s, a | \theta \right)}^{\text{calc. by Q-net}} \right)^{2}  \right]
\end{equation}

The full algorithm for the Deep Q-Learning with experience reply is described in more details in Appendix \ref{appendix_a1}. 

To evaluate the performances of the DQN algorithm, the researchers at Deep Mind took advantage of the Atari 2600 platform, offering 49 challenging games. The DQN algorithm outperformed the best existing reinforcement learning methods on 43 games without incorporating any prior knowledge about the Atari 2600 games. These outstanding results confirmed the superiority of the DQN algorithm and established it as a state-of-the-art technique in the Reinforcement Learning community.

\subsection{Double DQN (DDQN)}

The standard DQN algorithm is based on the Bellman equation, thus it includes maximization step as shown in Equation \ref{eq:loss_function}. For this reason, it learns unrealistically high action values, which tends to prefer overestimated to underestimated values.

In their paper \cite{van2016deep}, theoretically prove that the overestimations occur non-uniformly and negatively affect the performance of the DQN algorithm. Therefore, they proposed an extension of the standard DQN algorithm, called Double DQN (DDQN), in order to overcome these issues.

In the standard DQN algorithm, the decision for the next action is taken according to the following identity: $r + \gamma \max_{a^{\prime}} Q \left( s^{\prime}, a^{\prime}\right)$. For this reason, the same neural network is used to evaluate the Q-function and then to select the best action. 

In Double DQN, this process is decoupled. One neural network is used to evaluate the Q-function and a second neural network is used to select the best action. In mathematical notation, the next action is taken according to the following identity: $r + \gamma  Q \left( s^{\prime}, \argmax_{a^{\prime}} Q \left(s^{\prime}, a^{\prime}\right) \right)$.

\section{DQN for GO Chatbots}

The applications of the DQN algorithm are not only limited to the Atari 2600 games. Very recently the researchers started applying the Deep Q-Learning to various tasks, including the Goal-Oriented Dialogue Systems.

In the case of Goal-Oriented Chatbots, the agent is getting the new state $s_{t}$ from the Dialogue State Tracker (DST) and then it takes a new action $a_{t}$, based on the $\epsilon$-greedy policy. It means, with a probability $\epsilon \in \left[0,1\right]$ it will take a random action, while with a probability $1 - \epsilon$ it will take the action resulting with a maximal Q-value. We thus trade between the exploration and exploitation of the dialogue space. 
For each slot that might appear in the dialogue, the agent can take two actions: either to ask the user for a constraining value or to suggest to the user a value for that slot. Additionally, there is a fixed size of slot-independent actions, to open and close the conversation.

The agent receives positive and negative rewards accordingly, in order to force the agent to successfully conduct the dialogue. It is \textit{successful} if the number of totally required dialogue turns to reach the goal is less than a predefined maximal threshold $n_{max\_turns}$. For every additional dialogue turn, the agent receives a predefined negative reward $r_{ongoing}$. In the end, if the dialogue fails, it will receive a negative reward $r_{negative}$ equal to the negative of the predefined maximal allowed dialogue turns. If the dialogue is successful, it will receive a positive reward $r_{positive}$, two times the maximal allowed dialogue turns.

An important addition is the \textit{warm-starting} technique that fills the experience replay buffer with experiences coming from a successfully finished dialogues i.e. with positive experiences. This dramatically boosts the agent's performances before the actual training starts, as will be shown in Section \ref{sec:warmstart}. The training process continues with running a fixed number of independent training epochs $n_{epochs}$. In each epoch we simulate a predefined number of dialogues $n_{dialogues}$, thus filling the experience memory buffer. The result consists of mini-batches to train the underlying Deep Q-Net.

During the training process, when the agent reaches for the first time a success rate greater or equal to the success rate of a rule-based agent $s_{rule\_based}$, we flush the experience replay buffer, as described in detail in~\cite{li2017end}. We do this because the DQN-based agent cannot produce valuable experiences in the beginning, which are all stored in the experience buffer.

\section{Transfer Learning}

The main goal of this work is to study the impact of a widely used technique - \textit{Transfer Learning} on goal oriented bots. As the name suggests, transfer learning transfers knowledge from one neural network to another. The former is known as the source, while the latter is the target \cite{pan2010survey}. The goal of the transfer is to achieve better performance on the target domain with limited amount of training data, while benefiting from additional information from the source domain. In the case of dialogue systems, the input space for both source and target nets are their respective dialogue spaces. 

\begin{figure}[h!]
\centering
\begin{subfigure}[b]{.5\textwidth}
\centering
\includegraphics[width=\textwidth]{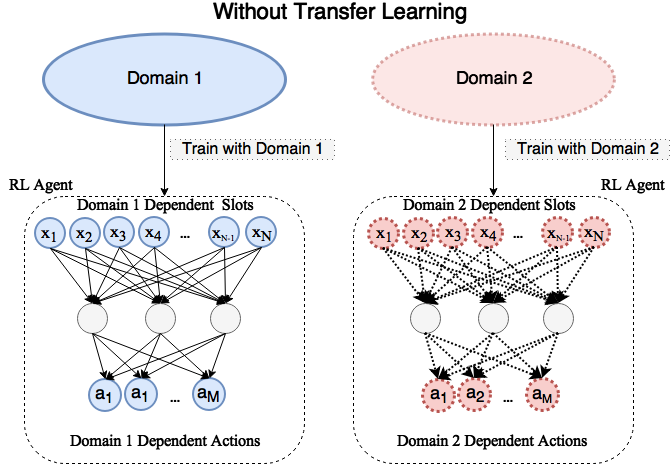}
\caption{}
\label{subfig:no_transfer_learning}
\end{subfigure}%
\begin{subfigure}[b]{.5\textwidth}
\centering
\includegraphics[width=\textwidth]{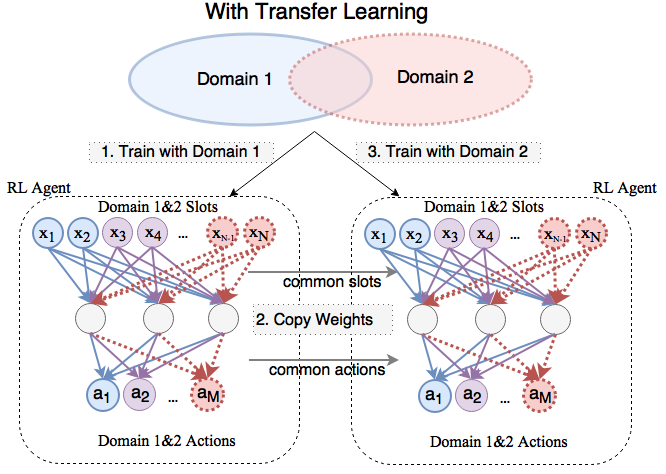}
\caption{}
\label{subfig:transfer_learning}
\end{subfigure}
\caption{Comparison of the Goal-Oriented Dialogue System training process, without transfer learning (left side) and with transfer learning (right side).}
\label{fig:no_transfer_learning_vs_transfer_learning}
\end{figure}

The training process without transfer learning, shown in Figure \ref{subfig:no_transfer_learning}, processes the two dialogue domains independently, starting from randomly initialized weights. The results are dialogue states from separate distributions. Additionally, the sets of actions the agents might take in each domain are also independent.

On the other hand, as depicted in Figure \ref{subfig:transfer_learning} if we want to benefit from transfer learning, we must model the dialogue state in both domains, as if they were coming from the same distribution. The sets of actions have to be shared too. The bots specialized in the source domain must be aware of the actions in the second domain, even if these actions are never used, and vice versa. This requirement stems from the impossibility of reusing the neural weights if the input and output spaces differ.
Consequently, when we train the model on the source domain, the state of the dialogue depends not only on the slots that are specific to the source, but also on those that only appear in the target one. This insight can be generalized to a plurality of source and target domains. The same holds for the set of actions.

\begin{algorithm}[h!]
\caption{Transfer Learning Pseudocode}
\label{alg:transfer_learning}

\begin{algorithmic}[1]
\Procedure{InitializeWeights}{sourceWeights, commonSlotIndices, commonActionIndices}

\State $ targetWeigths \gets \textit{RandInit()} $

\For{$i$ in $\textit{commonSlotIndices}$}
   \State $ \textit{targetWeigths} \left[ i \right] \gets sourceWeights \left[ i \right] $
\EndFor

\For{$i$ in $\textit{commonActionIndices}$}
   \State $ \textit{targetWeigths} \left[ i \right] \gets sourceWeights\left[ i\right ]$
\EndFor

\State \textbf{return} \textit{targetWeigths}
\EndProcedure
\end{algorithmic}

\end{algorithm}

When training the target domain model, we no longer randomly initializing all weights. The weights related to the source domain - both for slots and actions - are copied from the source model. The pseudocode for this weight initialization is portrayed in the Algorithm \ref{alg:transfer_learning}.

\chapter{Experiments and Results}
\label{chap:experiments_results}

In this chapter we present all valuable experiments and the obtained results. Our work is based on the work from \cite{li2017end}. For this reason, in the first part we briefly present the baseline experiments and results. In the second part we focus on the transfer learning experiments.

\section{Baseline Experiments}

\begin{figure}[b!]
\centering
\begin{subfigure}{0.5\textwidth}
\centering
\includegraphics[width=0.98\textwidth]{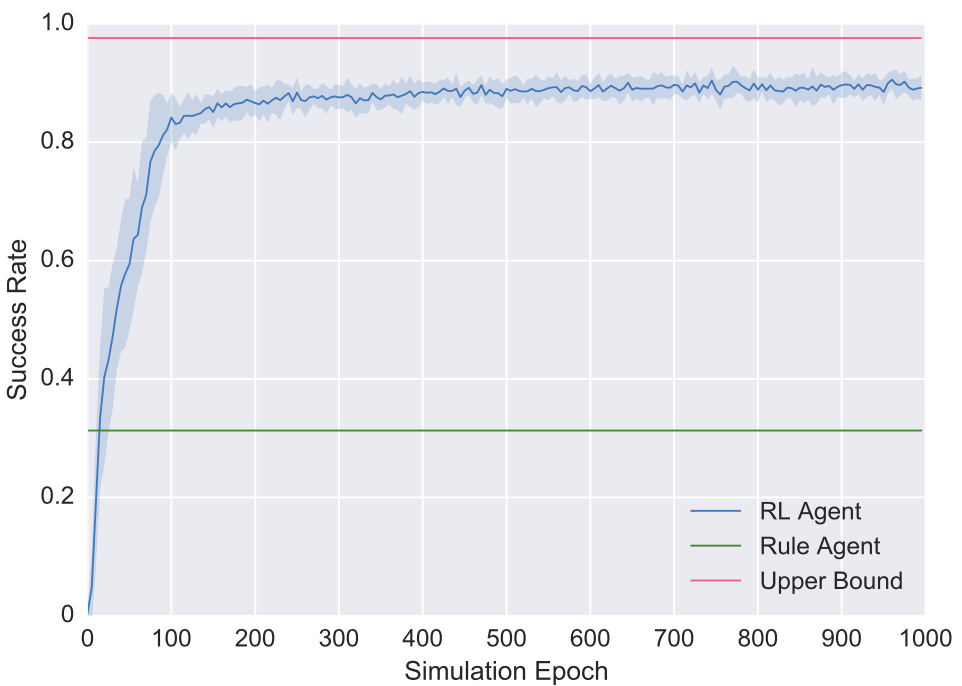}
\caption{Learning curve on a semantic level}
\label{subfig:frame_level_base}
\end{subfigure}%
\begin{subfigure}{0.5\textwidth}
\centering
\includegraphics[width=0.99\textwidth]{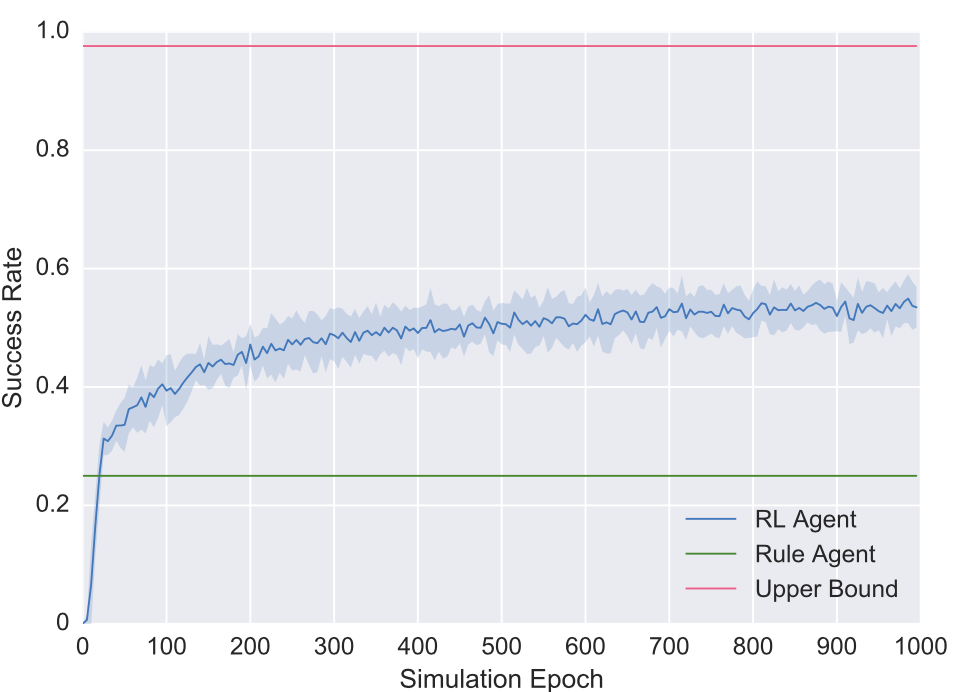}
\caption{Learning curve on a natural language level}
\label{subfig:nl_level_base}
\end{subfigure}
\caption{Baseline experimental results for 100 runs with a 95\% confidence interval}
\label{fig:baseline_exp}
\end{figure}

In all baseline experiments, the Chatbot is trained on the \textit{Movie Booking} domain. The type of slots for this domain are given in Figure \ref{fig:slot_type}. The size of the training set is 128 user goals. The maximal number of allowed dialogue turns is set to $n_{max\_turns} = 20$, thus the negative reward for a failed dialogue is $r_{negative} = -20$, while the positive reward for a successful dialogue is $r_{positive} = 40$. In all experiments we use a warm-starting and train for $n_{epochs} = 1000$ epochs, each simulating $n_{dialogues} = 100$ dialogues. This is a bit huge number of epochs to run and we believe that the chatbot is overfitting. However, our intention was to reproduce the results from the baseline paper.

We present the results for both levels: semantic level and natural-language level. In Figure \ref{subfig:frame_level_base} the learning curve for the training on semantic level is shown, while in Figure \ref{subfig:nl_level_base} the learning curve for the training on natural language level is shown. The same experiment is repeated 100 times, thus the results are reported with a 95\% confidence interval. Due to the noise introduced by the NLU and NLG unit, the Chatbot performance on natural language level is considerably lower than the performance on the semantic level.

\section{Transfer Learning Experiments}

In this set of experiments, we operate on the semantic level, removing the noise introduced by the NLU and NLG units. We want to focus exclusively on the impact of transfer learning techniques on dialog management. The details of the system implementation\footnote{Link to the GitHub repository: \href{https://github.com/IlievskiV/Master_Thesis_GO_Chatbots}{https://github.com/IlievskiV/Master\_Thesis\_GO\_Chatbots}} are presented in Appendix \ref{appenix_a3}.

\subsection{Setup of Experiments}

All experiments are executed using a setup template. Firstly, we train a model on the \textit{source domain} and reuse the common knowledge to boost the training and testing performance of the model trained on a different, but similar \textit{target domain}. Secondly, we train a model exclusively on the target domain, without any prior knowledge. This serves as a baseline. Finally, we compare the results of these two models. We thus have two different cases:

\begin{enumerate}
	\item \textit{Domain Overlap} - the source \(Movie Booking\) and target \(Restaurant Booking\) domains are different, but share a fraction of the slots.
    \item \textit{Domain Extension} - the source domain, now \(Restaurant Booking\), is extended to \textit{Tourist Information}, that  contains all the slots from the source domain along with some additional ones.
\end{enumerate}

\begin{figure}[h!]
\centering
\includegraphics[width=.75\textwidth]{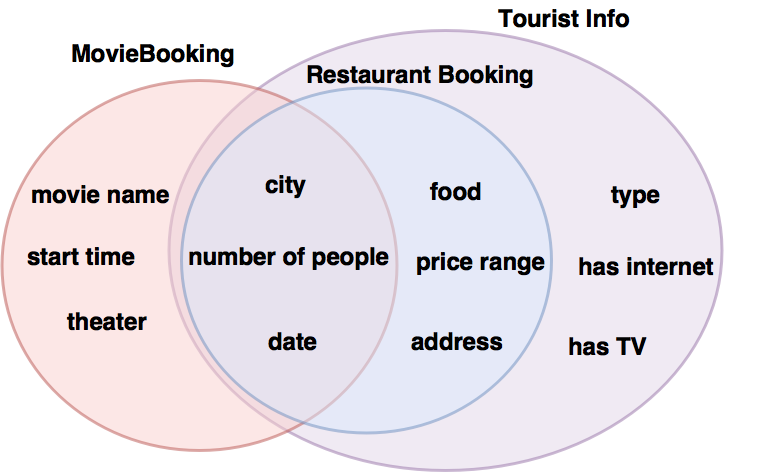}
\caption{Slot types in the three different domains}
\label{fig:slot_type}
\end{figure}

The reason for the choice of source domain in the domain overlap case is designed to enable a direct comparison to the results of ~\cite{li2017end}, who built a GO bot for movie booking. For the domain extension case, the only combination available was \(Restaurant - Tourism\).
The type of slots in each domain are given in Figure \ref{fig:slot_type}. For each domain, we have a training set of 120 user goals, and a testing set of 32 user goals.

Following the above mentioned setup template, we conduct two sets of experiments for each of the two cases. The first set shows the overall performance of the models leveraging the transfer learning approach. The second set shows the effects of the warm-starting jointly used with the transfer learning technique.

In all experiments, when we use a warm-starting, the criterion is to fill agent's buffer, such that 30 percent of the buffer is filled with positive experiences (coming from a successful dialogue). After that, we train for $n_{epochs} = 50$ epochs, each simulating $n_{dialogues} = 100$ dialogues. We flush the agent's buffer when the agent reaches, for a first time, a success rate of $s_{rule\_based} = 0.3$. We set the maximal number of allowed dialogue turns $n_{max\_turns}$ to 20, thus the negative reward $r_{negative}$ for a failed dialogue is $-20$, while the positive reward $r_{positive}$ for a successful dialogue is $40$. In the consecutive dialogue turns over the course of the conversation, the agent receives a negative reward of $r_{ongoing} = -1$. In all cases we set $\epsilon = 0.05$ to leave a space for exploration. By using this hyperparameters, we prevent the system to overfit and to generalize very well over the dialogue space. Finally, we report the success rate as a performance measure.

\subsection{Training with Less Data}

Due to labeling costs, the availability of in-domain data is the bottleneck for training successful and high performing Goal-Oriented chatbots. We thus study the effect of transfer learning on training bots in data-constrained environments.

From the available 120 user goals for each domain's training set, we randomly select subsets of 5, 10, 20, 30, 50 and all 120. We then warm-start and train both the independent and transfer learning models on these sets. We test the performance on both the training set (\textit{training performance}) and the full set of 32 test user goals (\textit{testing performance}). We repeat the same experiment 100 times, in order to reduce the uncertainty introduced by the random selection. Finally, we report the success rate over the user goal portions with a 95\% confidence interval.

\begin{figure}[h!]
\centering
\begin{subfigure}[t]{\textwidth}
\centering
\includegraphics[width=.49\textwidth]{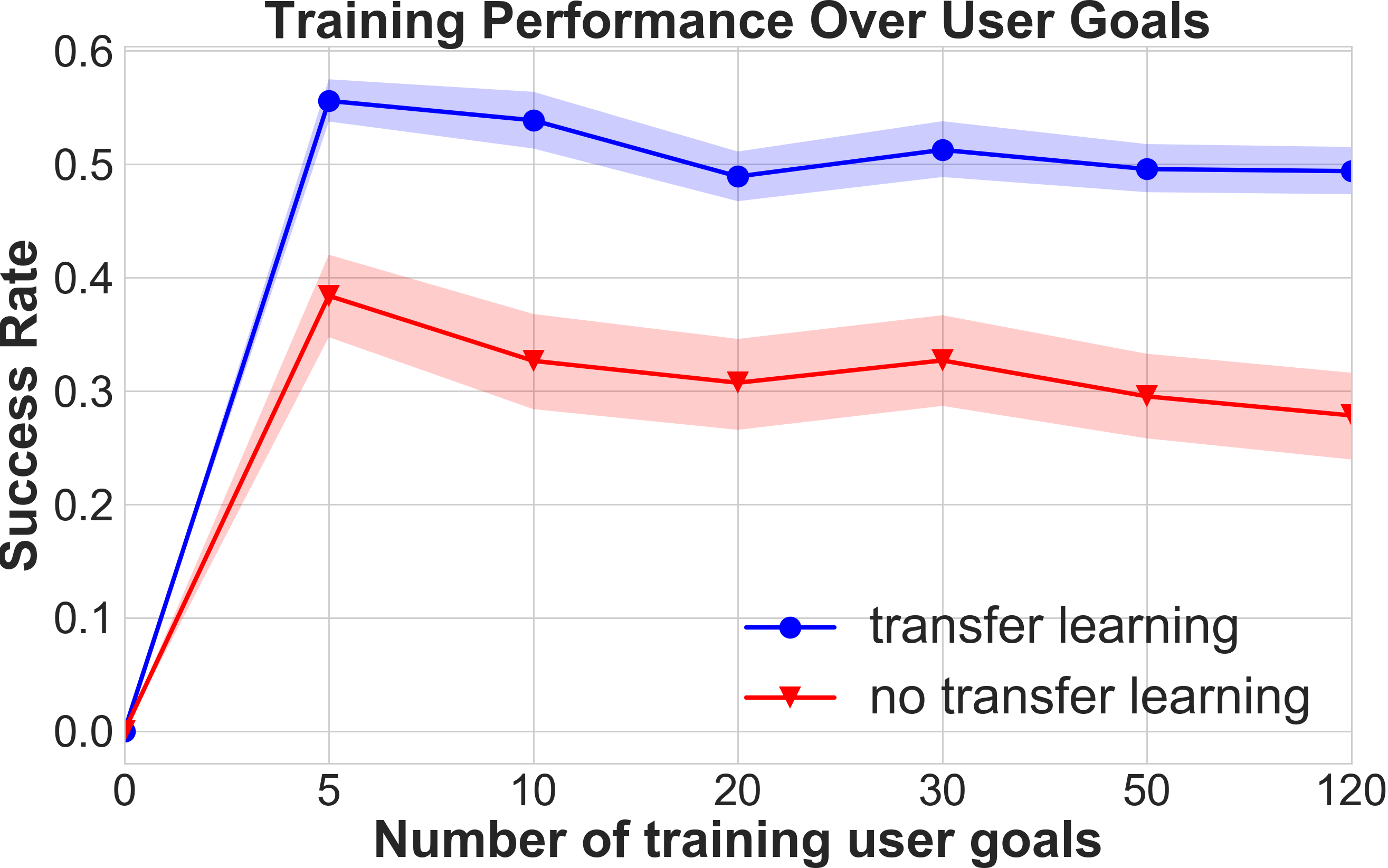}
\includegraphics[width=.49\textwidth]{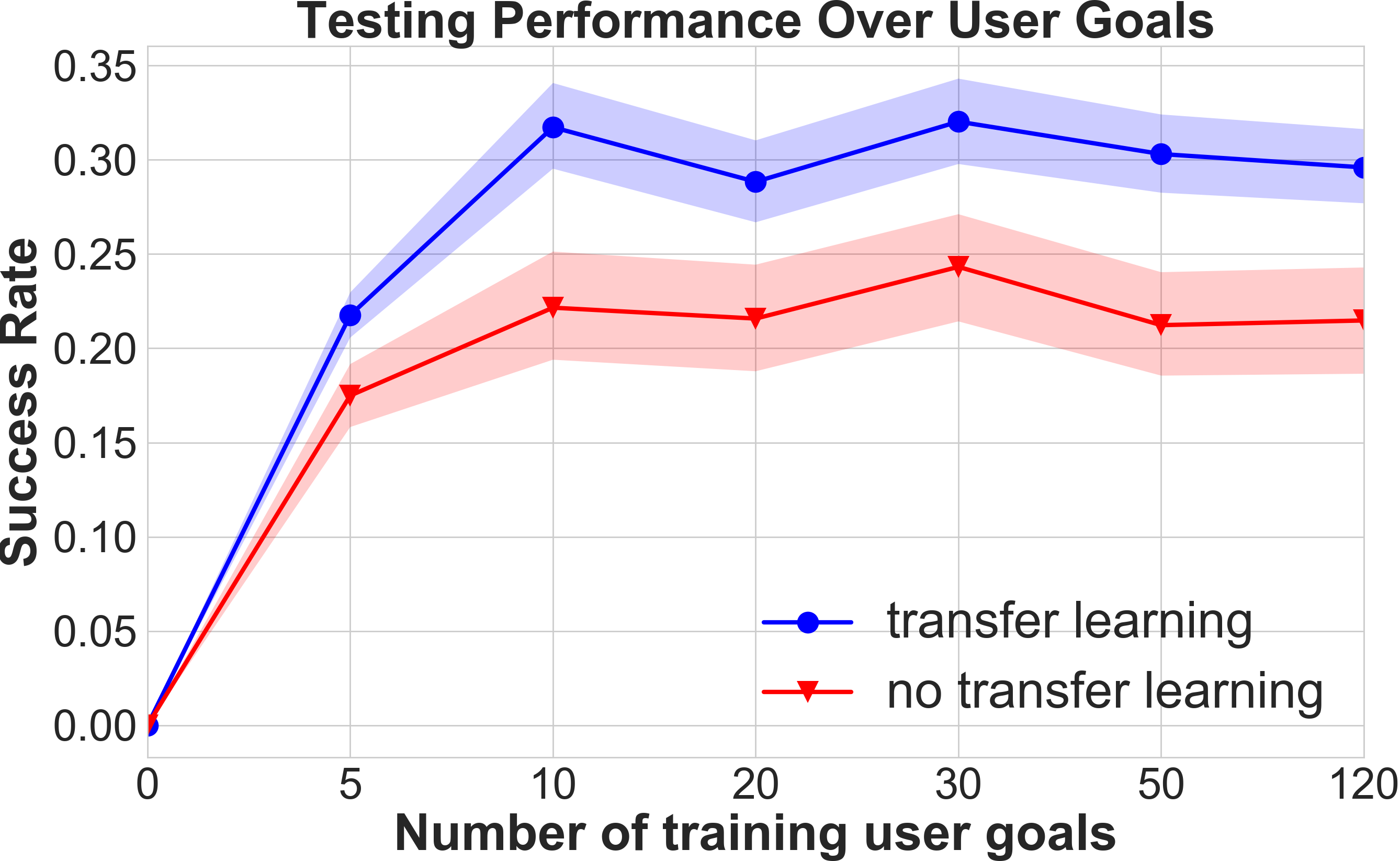}
\caption{Restaurant Booking with pre-training on Movie Booking domain}
\label{subfig:warm_up_rest_booking}
\end{subfigure}\\%
\begin{subfigure}[b]{\textwidth}
\centering
\includegraphics[width=.49\textwidth]{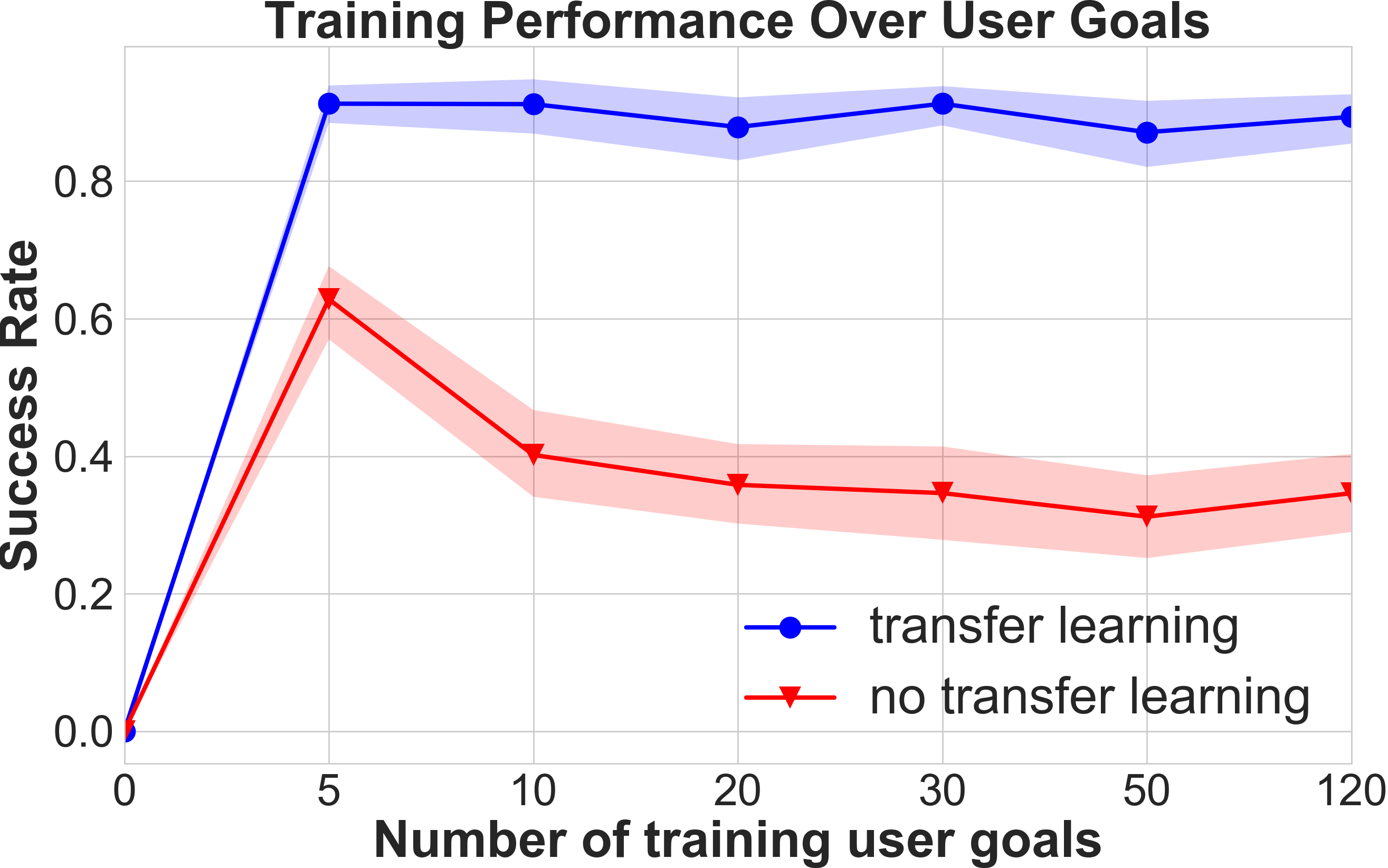}
\includegraphics[width=.5\textwidth]{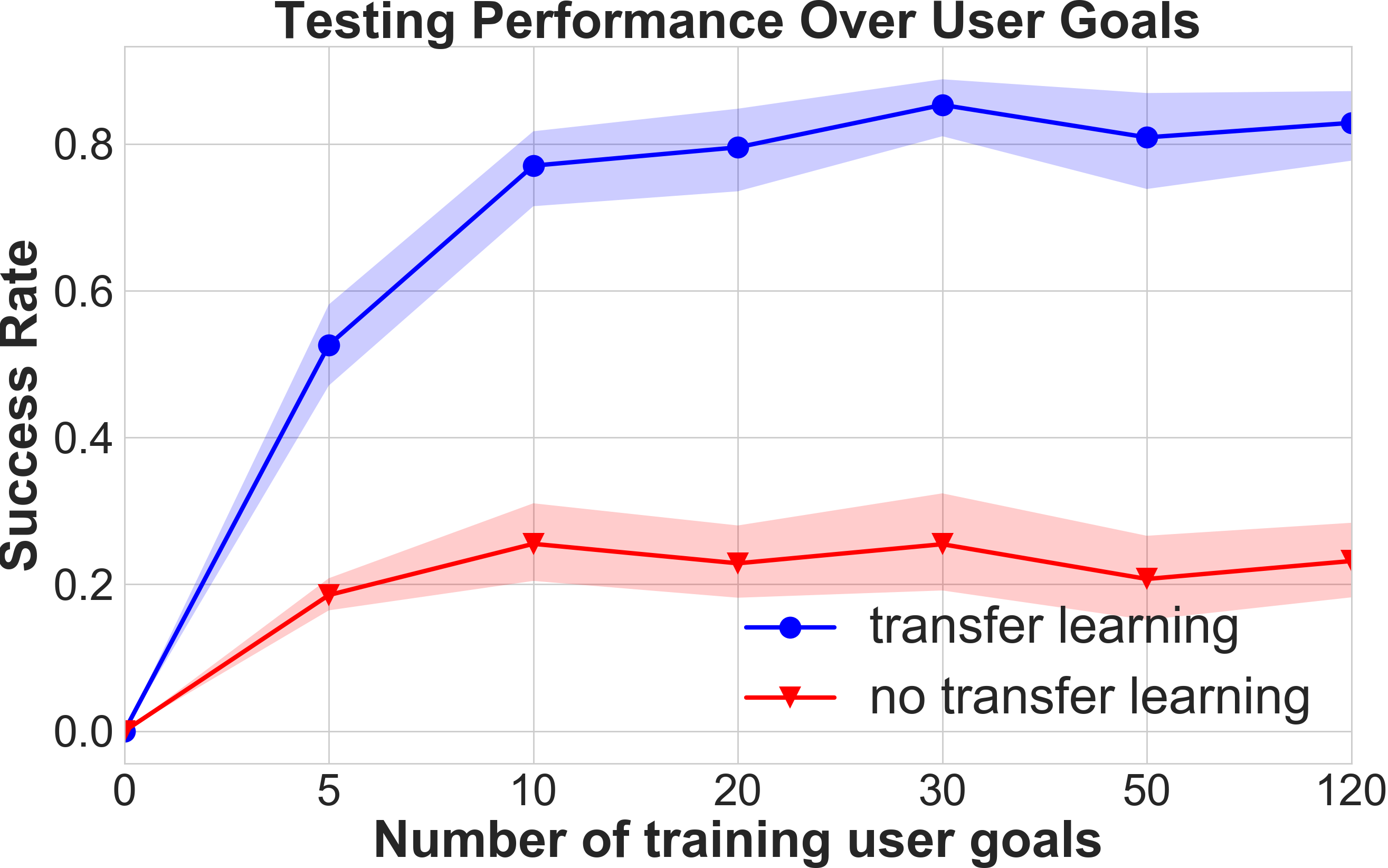}
\caption{Tourist Info with pre-training on Restaurant Booking domain}
\label{subfig:warm_up_tourist_info}
\end{subfigure}
\caption{Average training and testing success rates with 95\% confidence, for 100 runs over a randomly selected user goal portions of size 5, 10, 20, 30, 50 and 120, for both models: with and without transfer learning.}
\label{fig:warming_up_user_goal_portions}
\end{figure}

The training and testing results, in the first case of domain overlapping, are shown in Figure \ref{subfig:warm_up_rest_booking}. The success rate of the model obtained with transfer learning is 65\% higher than that of the model trained without any external prior knowledge. In absolute terms the success rate climbs on average from 30\% to 50\%. For the test dataset, transfer learning improves the success rate from 25\% to 30\%, for a still noteworthy 20\% relative improvement.

In the case of domain extension, the difference between the success rates of the two models is even larger (Figure \ref{subfig:warm_up_tourist_info}). This was expected, as the extended target domain contains all slots from the source domain, therefore not losing any source domain information. The overall relative success rate boost for all user goal portions is on average 112\%, i.e. a move from 40\% to 85\% in absolute terms. For the test set, this difference is even larger, from 22 to 80\% absolute success rate, resulting in 263\% relative boost. 

These results show that transferring the knowledge from the source domain, we boost the target domain performance in data constrained regimes.

\subsection{Faster Learning}
\label{sec:warmstart}

In a second round of experiments, we study the effects of the transfer learning in the absence and in combination with the warm-starting phase. 
As warm starting requires additional labeled data, removing it further reduces the amount of labeled data needed. We also show that the two methods are compatible, leading to very good joint results.

\begin{figure}[h!]
\centering
\begin{subfigure}[t]{\textwidth}
\centering
\includegraphics[width=.49\textwidth]{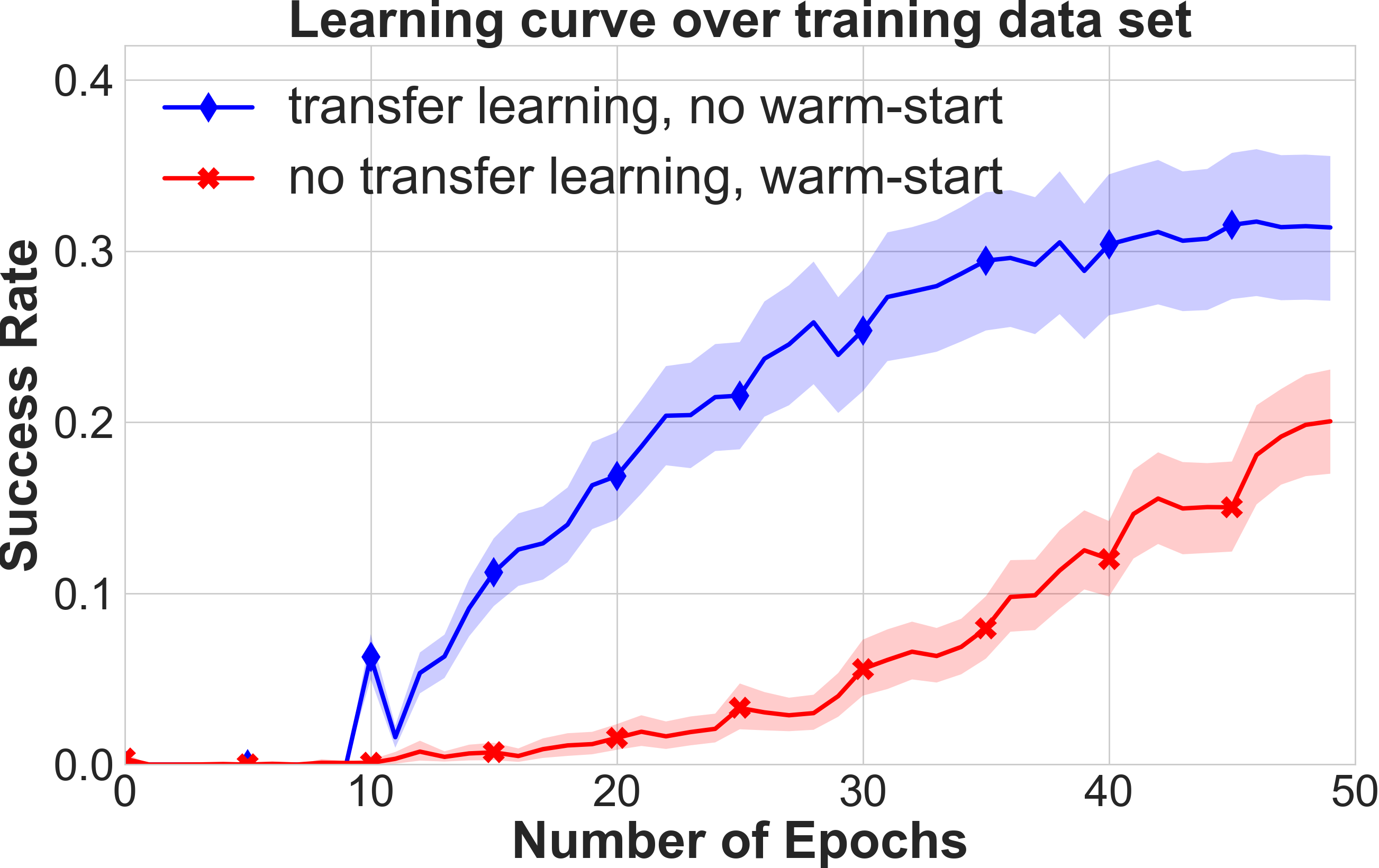}
\includegraphics[width=.49\textwidth]{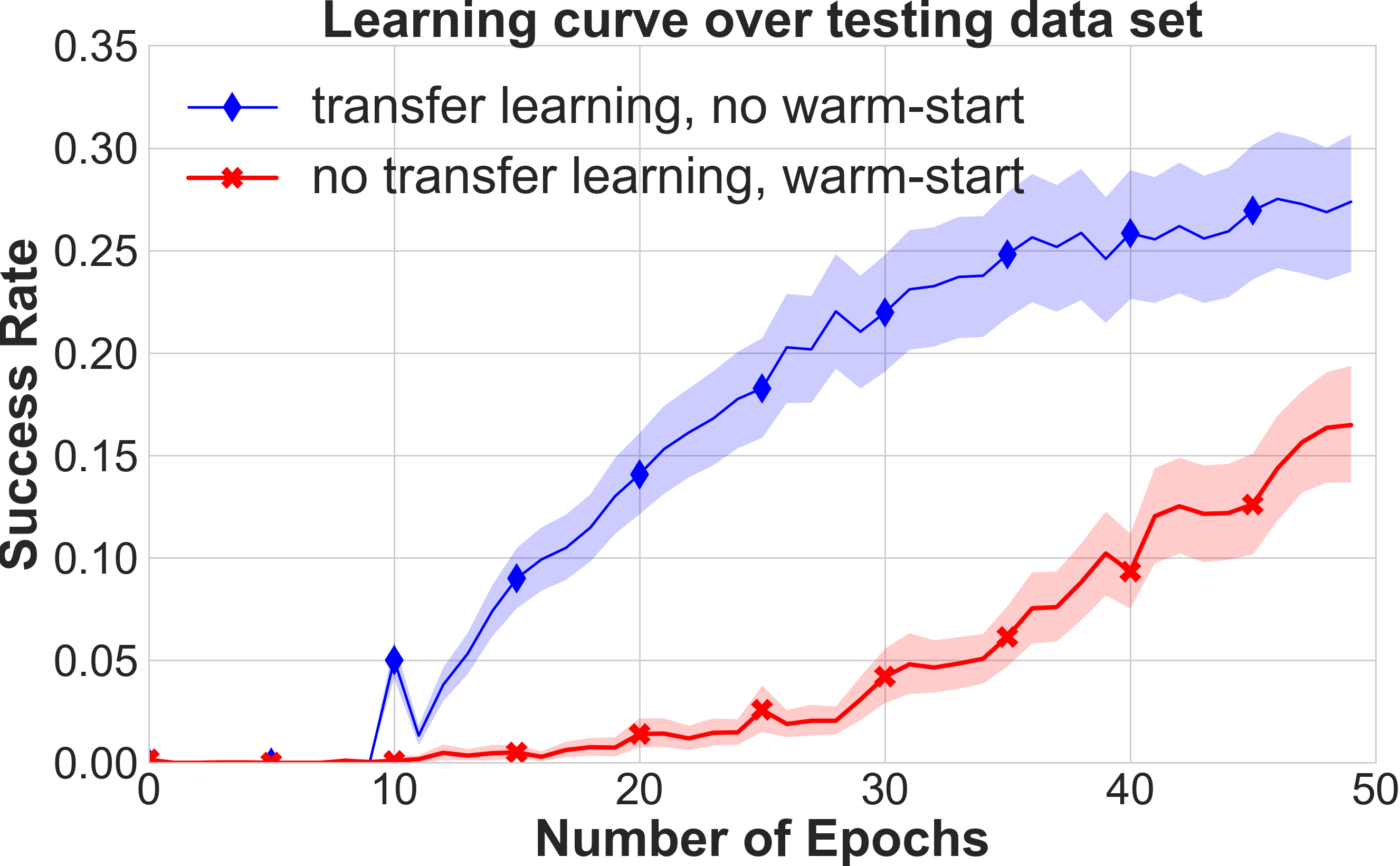}
\caption{Restaurant Booking with pre-training on Movie Booking domain}
\label{subfig:no_warm_up_rest_booking}
\end{subfigure}\\%
\begin{subfigure}[b]{\textwidth}
\centering
\includegraphics[width=.49\textwidth]{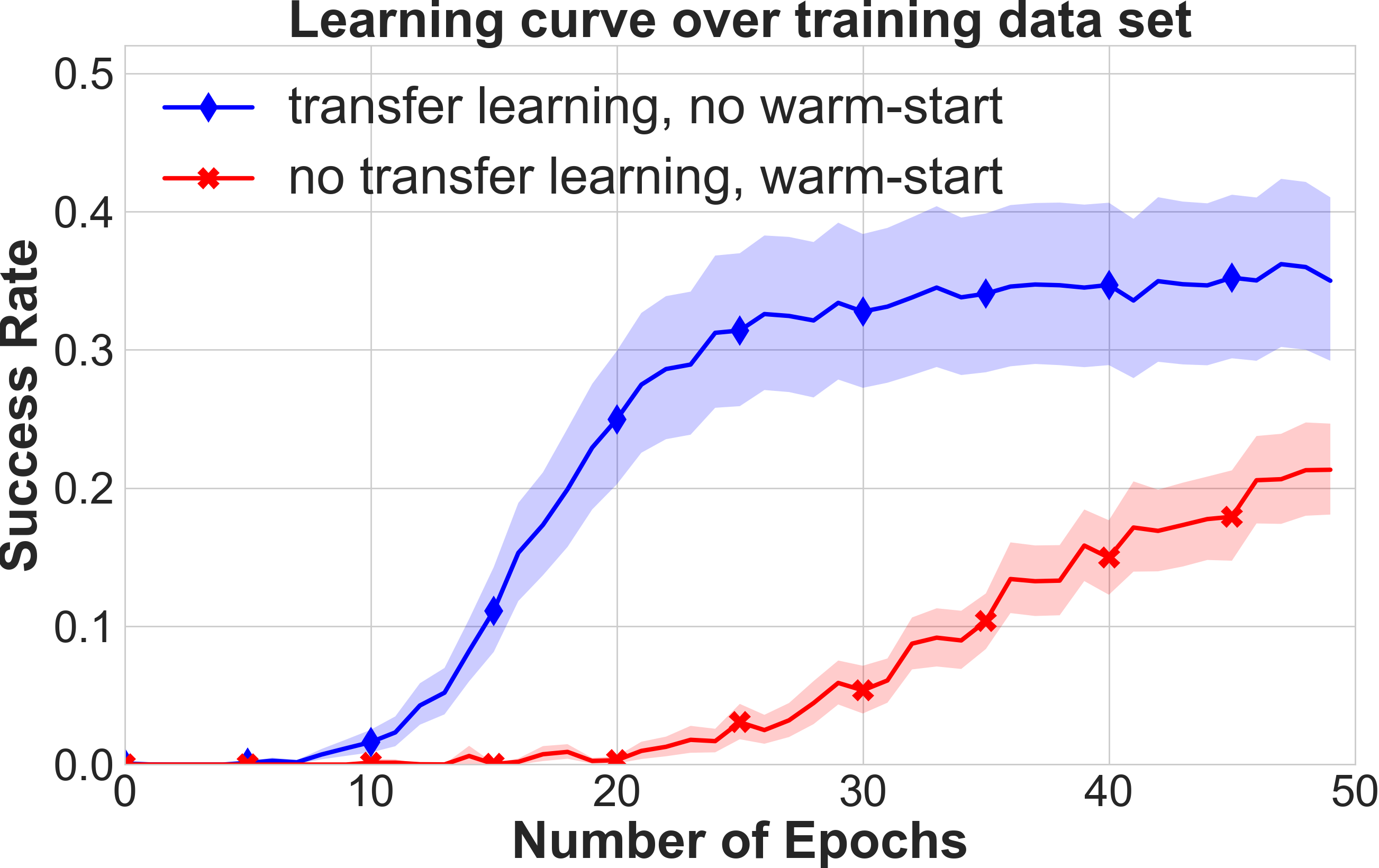}
\includegraphics[width=.5\textwidth]{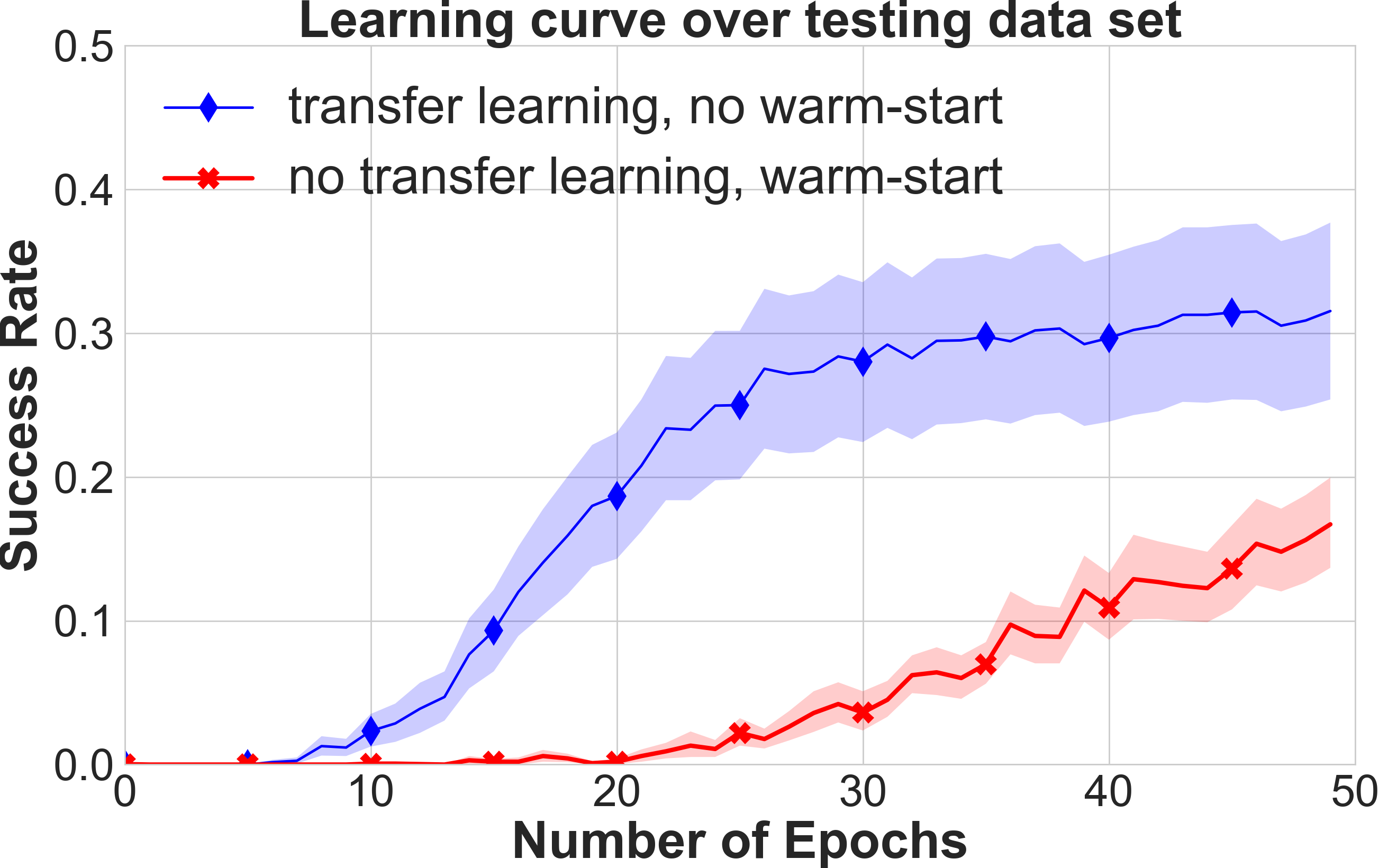}
\caption{Tourist Info with pre-training on Restaurant Booking domain}
\label{subfig:no_warm_up_tourist_info}
\end{subfigure}
\caption{Average training and testing success rates with 95\% confidence, for 100 runs over the number of epochs, for both models: with and without transfer learning. The model with transfer learning is not warm-started.}
\label{fig:no_warming_up_learning_curve}
\end{figure}

We report the training and testing learning curves (success rate over the number of training epochs), such that we use the full dataset of 120 training user goals and the test set of 32 user goals. We repeat the same process 100 times and report the results with a 95\% confidence interval.
The performances in the first case of domain overlapping are shown in Figure \ref{subfig:no_warm_up_rest_booking}, while for the other case of domain extension, in Figure \ref{subfig:no_warm_up_tourist_info}. The bot using transfer learning, but no warm-starting, shows better learning performances than the warm-started model without transfer learning. Transfer learning is thus a viable alternative to warm starting.

\begin{figure}[h!]
\centering
\begin{subfigure}[t]{\textwidth}
\centering
\includegraphics[width=.5\textwidth]{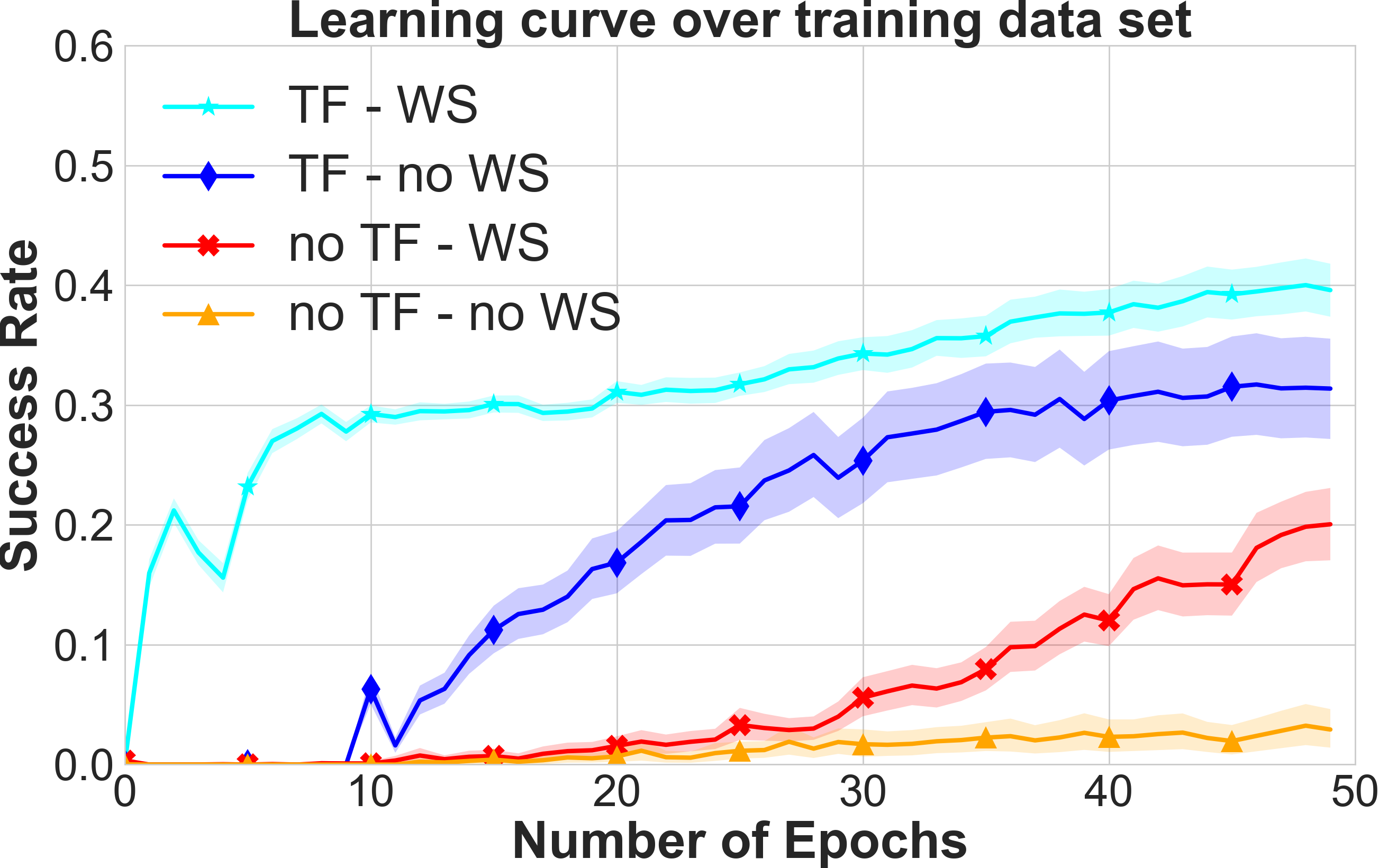}
\includegraphics[width=.49\textwidth]{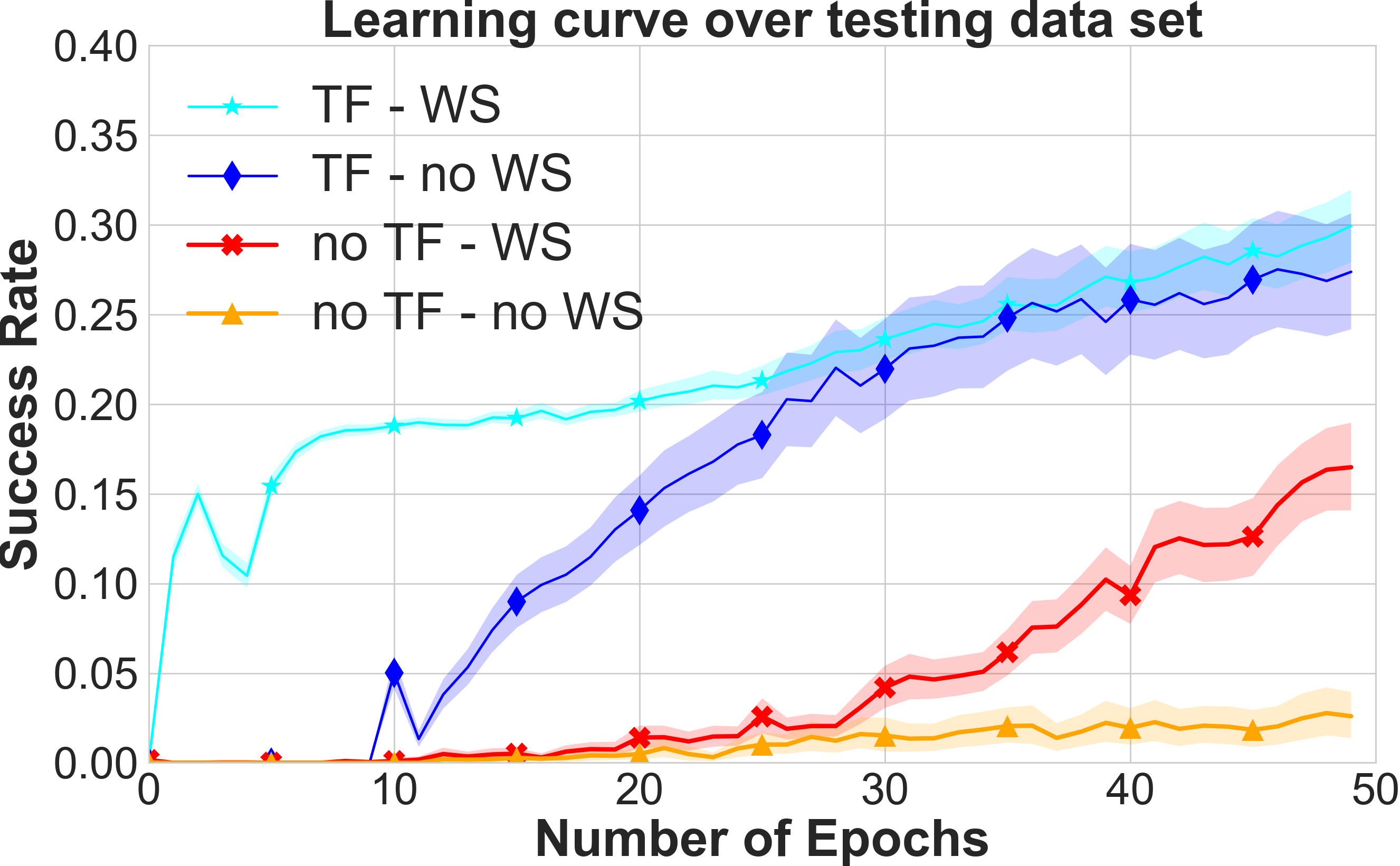}
\caption{Restaurant Booking with pre-training on Movie Booking domain}
\label{subfig:all_cases_rest_booking}
\end{subfigure}\\%
\begin{subfigure}[b]{\textwidth}
\centering
\includegraphics[width=.49\textwidth]{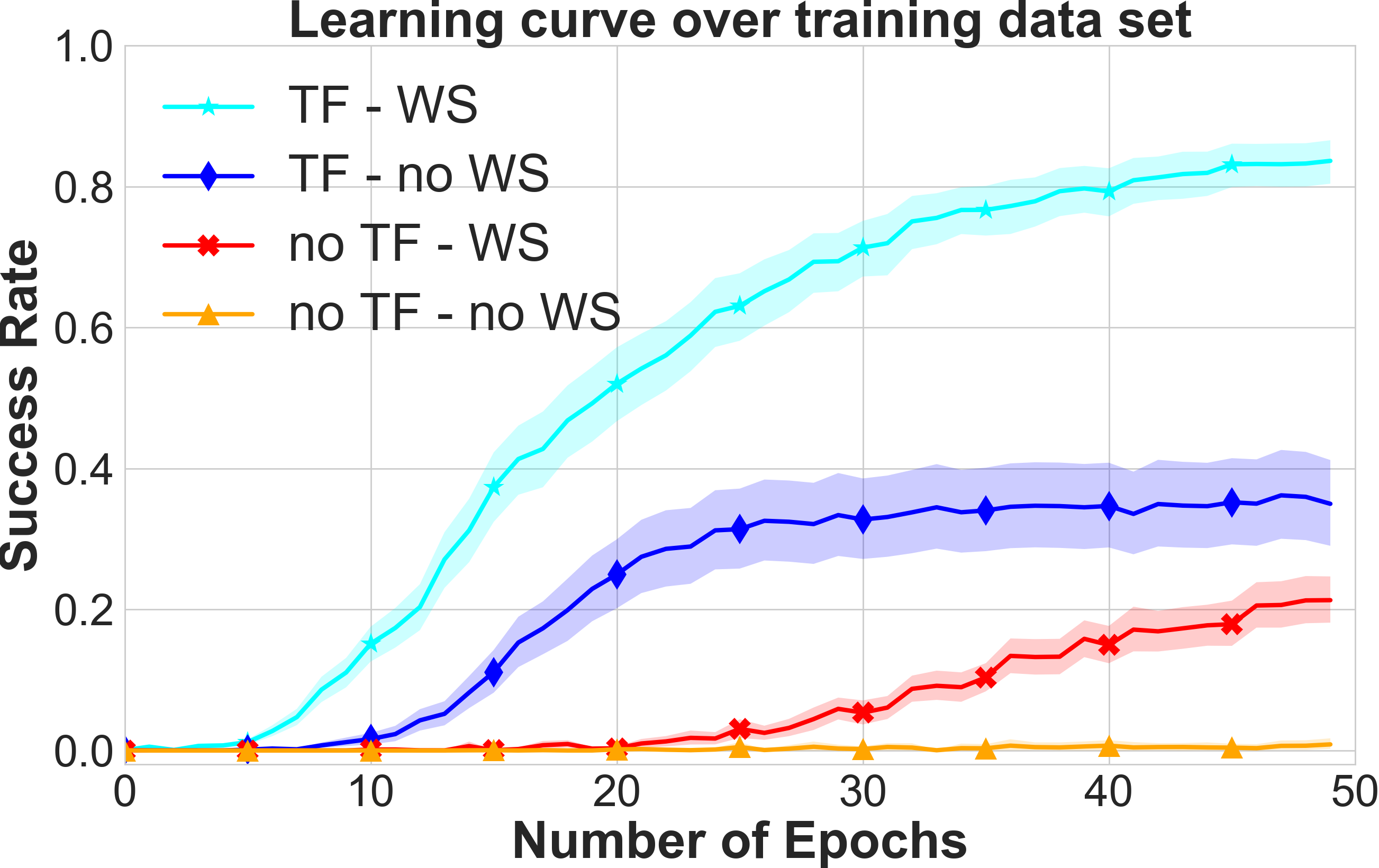}
\includegraphics[width=.5\textwidth]{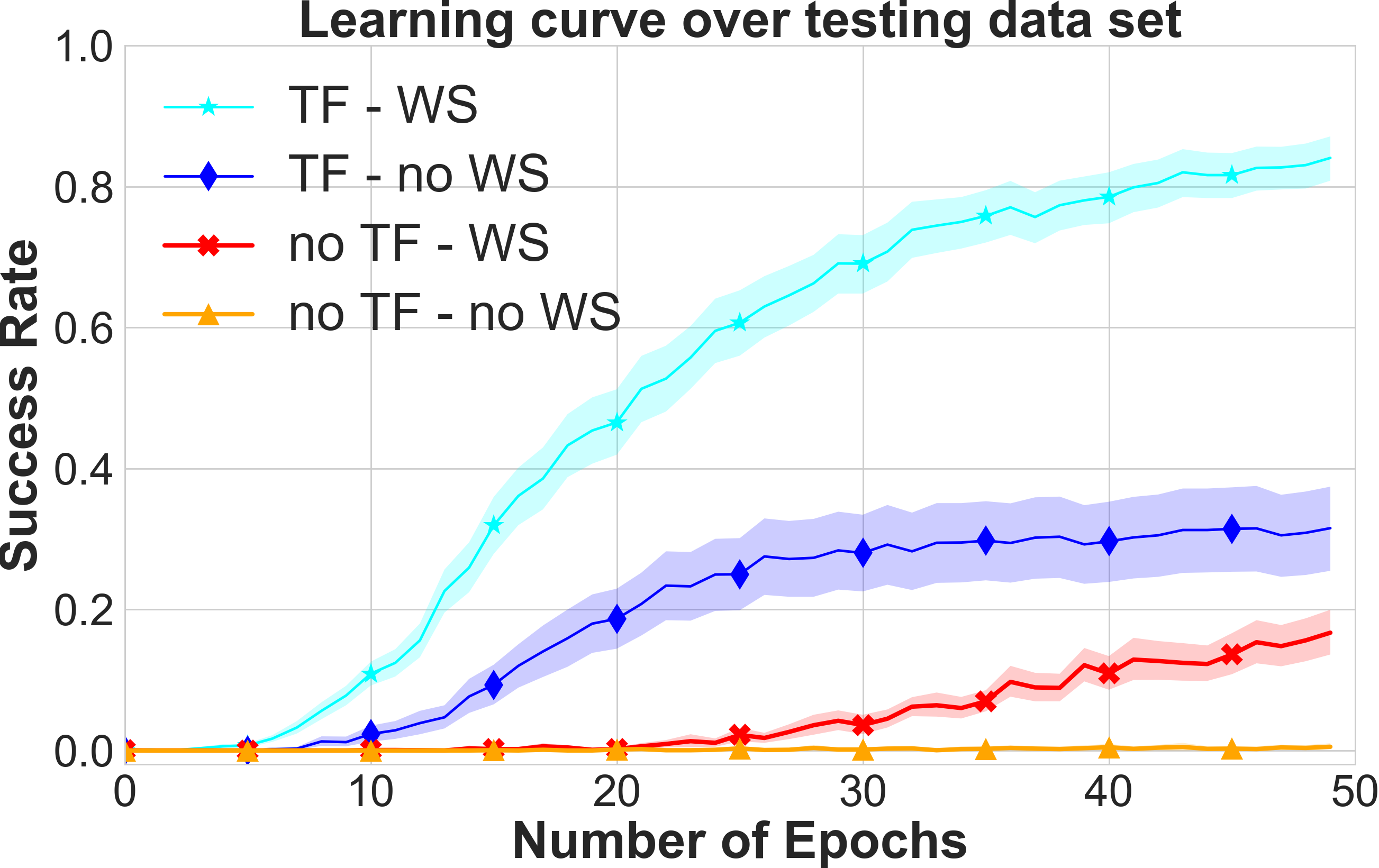}
\caption{Tourist Info with pre-training on Restaurant Booking domain}
\label{subfig:all_cases_tourist_info}
\end{subfigure}
\caption{Success rates for all model combinations - with and without Transfer Learning (TF), with and without Warm Starting (WS).}
\label{fig:all_cases_learning_curve}
\end{figure}

However, models based on transfer learning have a significant variance, as the learning is progressing. This happens because in many experiment runs the success rate over all epochs is 0. In those cases, the agent does not find an optimal way to learn the policy in the early stages of the training process. This results with filling its experience replay buffer mostly with negative experiences. Consequently, in the later stages, the agent is not able to recover. This makes a combination with warm starting desirable.

For convenience reasons, in Figure \ref{fig:all_cases_learning_curve} we show all possible cases of using and combining the transfer learning and warm-starting techniques. We can see that the model combines the two techniques performs the best by a wide margin. This leads to a conclusion that the transfer learning is complimentary to the warm-starting, such that their joint application brings the best outcomes.
\chapter{Conclusion and Future Work}
\label{chap:conclusion}

In this thesis, we examined in depth the Goal-Oriented Chatbots, especially the Reinforcement Learning-based. We show that the \textit{Transfer Learning} technique can be successfully applied to boost the performances of the RL-based Goal-Oriented Chatbots. We do this for two different use cases: $i)$ when the source and the target domain overlap, and $ii)$ when the target domain is an extension of the source domain. 

We show the advantages of the transfer learning in a low data regime for both cases. When a low number of user goals is available for training in the target domain, transfer learning makes up for the missing data. Even when the whole target domain training data is available, the transfer learning benefits are maintained, with the success rate increasing threefold.

We also demonstrate that the transfer knowledge can be a replacement of the warm-starting period in the agents or can be combined with it for best results. 

Last but not the least, we create and share two datasets for training Goal-Oriented Dialogue Systems in the domains of Restaurant Booking and Tourist Information.

With the promising results we achieved during the work on this thesis, the following possibilities are worth trying to investigate into:
\begin{itemize}
\item Study the effects of the \textit{Transfer Learning} when the Chatbot is working on a natural language level.
\item Build better \textit{User Simulators}, since the simulator in this work is hand-crafted and too limited.
\item To build better and more robust Dialogue State Trackers.
\end{itemize}

\appendix
\chapter{Appendix}
\label{appendix_a}

\section{Standard DQN algorithm}
\label{appendix_a1}

\begin{algorithm}[h!]
\caption{Deep Q-Learning with Experience Replay \cite{mnih2015human}}
\label{alg:dqn_algo}

\begin{algorithmic}[1]

\Procedure{DQN}{N, M, T, $\epsilon$, $\gamma$}

\State Initialize replay memory $\mathcal{D}$ to capacity $N$
\State Initialize action-value function $Q$ with random weights $\theta$

\For{$episode$ in $1,M$}
\State Initialize sequence $s_{1}={x_{1}}$ and preprocessed sequence $\phi_{1} = \phi(s_{1})$

\For{$t$ in $1,T$}
    \State With probability $\epsilon$ take a random action $a_{t}$,
    \State otherwise select $a_{t} = \max_{a}Q^{\star}(\phi(s_{t}), a; \theta)$
    \State Execute  action $a_{t}$ and observe a reward $r_{t}$ and a new state $x_{t+1}$
    \State Set $s_{t+1} = s_{t}, a_{t}, x_{t+1}$ and preprocess $\phi_{t+1} = \phi(x_{t+1})$
    \State Store transition $(\phi_{t}, a_{t}, r_{t}, \phi_{t+1})$ in $\mathcal{D}$
    \State Sample random minibatch of transitions $(\phi_{j}, a_{j}, r_{j}, \phi_{j+1})$ from $\mathcal{D}$
    \State Set $y_{j} = \begin{cases} 
      r_{j} \quad \qquad \qquad \qquad \qquad \qquad \qquad \text{for terminal $\phi_{j+1}$} \\
      r_{j} + \gamma \max_{a^{\prime}}Q^{\star}(\phi_{j+1},       a^{\prime}; \theta) \qquad \text{for non-terminal $\phi_{j+1}$}
    \end{cases}$
    
    \State Perform a gradient descent step on $(y_{j} - Q(\phi_{j}, a_{j};\theta))^{2}$ (Eq. \ref{eq:loss_function})

\EndFor
\EndFor

\EndProcedure
\end{algorithmic}

\end{algorithm}

\section{System Implementation}
\label{appenix_a3}

\begin{figure}[h!]
 \centering 
 \includegraphics[width=\textwidth]{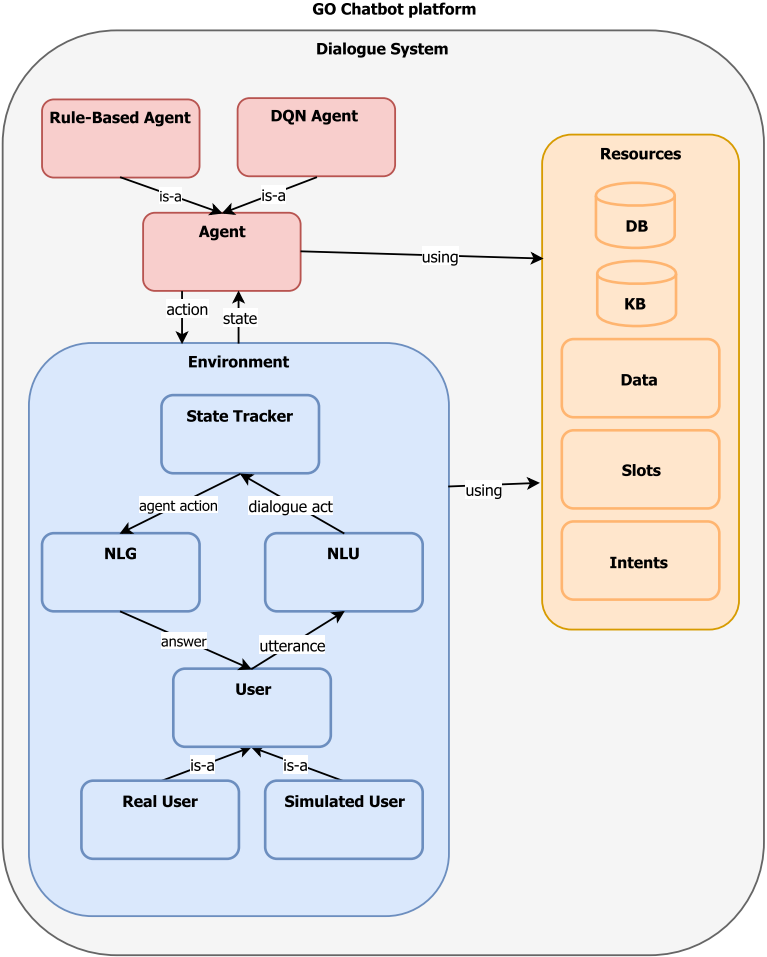}
 \caption{High-level overview of the system for training Goal-Oriented Chatbots}
 \label{fig:GO_platform}
\end{figure}

\begin{figure}[h!]
 \centering 
 \includegraphics[angle=90, origin=c, width=\textwidth, height=170mm]{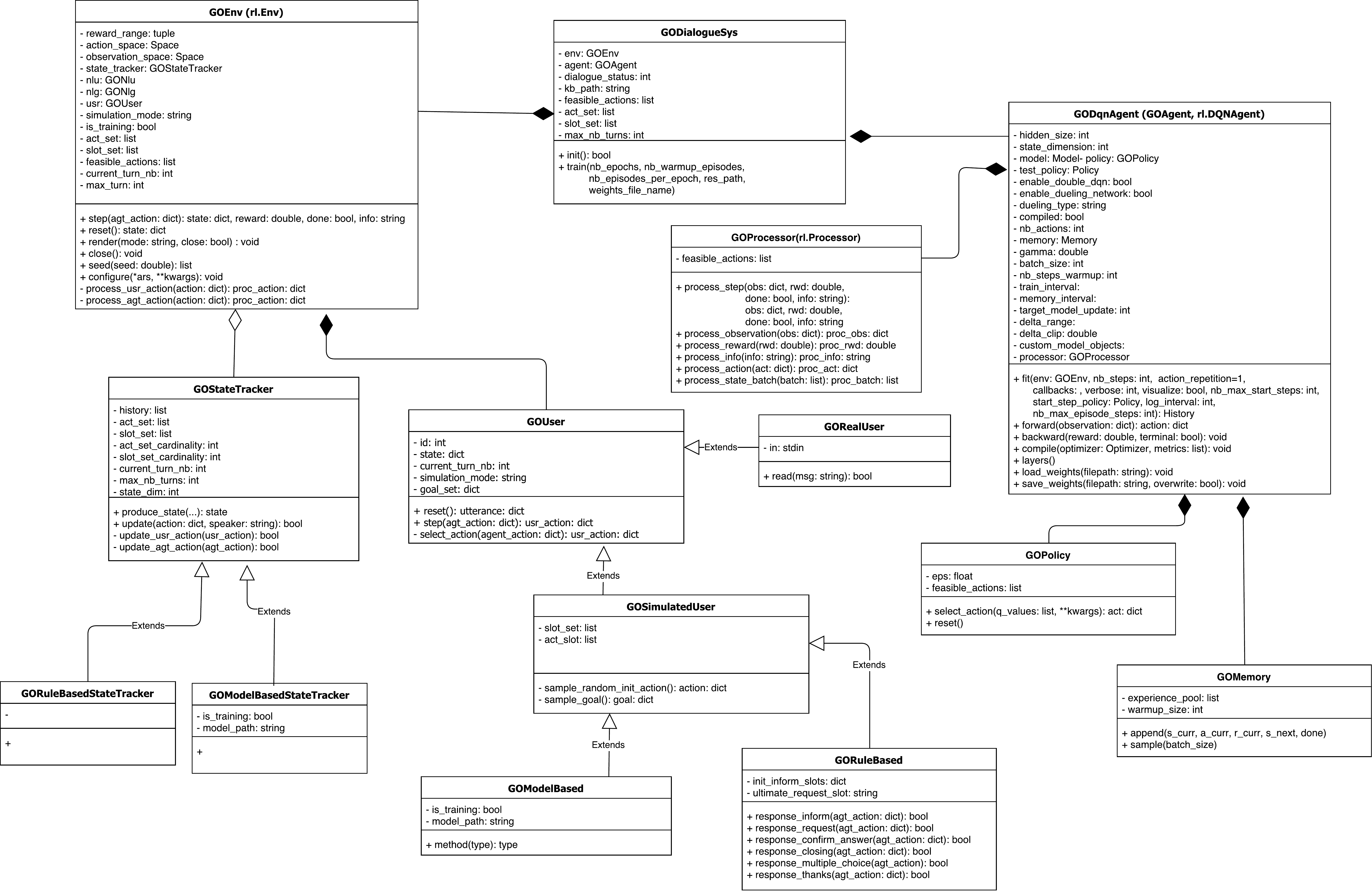}
 \caption{UML Diagram of the system for training Goal-Oriented Chatbots}
 \label{fig:uml_diagram}
\end{figure}

\begin{figure}[h!]
 \centering 
 \includegraphics[width=\textwidth]{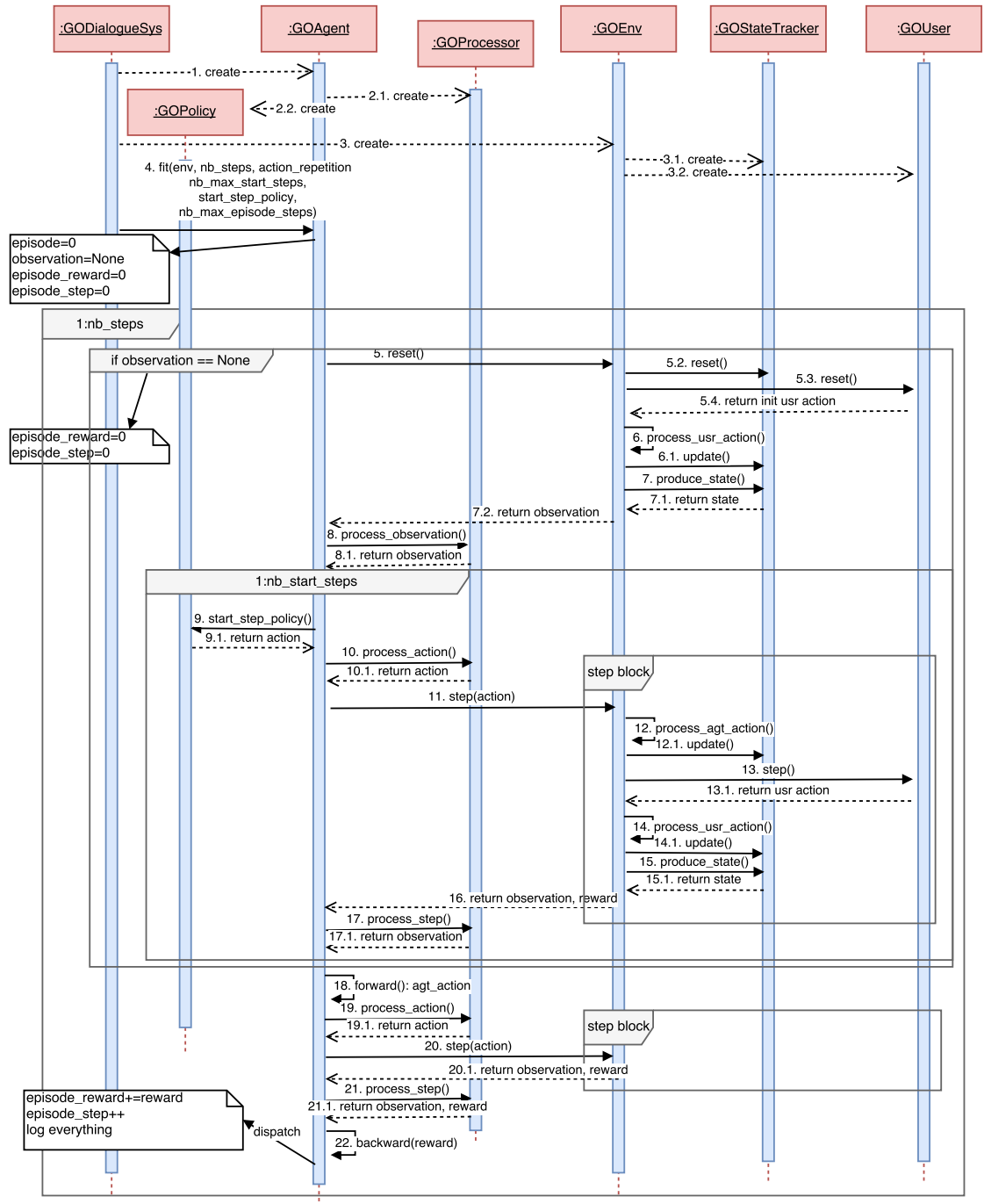}
 \caption{Sequence Diagram of the system for training Goal-Oriented Chatbots}
 \label{fig:sequence_diagram}
\end{figure}
\bibliographystyle{apalike}
\bibliography{tail/bibliography}

\begin{thebibliography}{}

\bibitem[Banchs, 2012]{banchs2012movie}
Banchs, R.~E. (2012).
\newblock Movie-dic: a movie dialogue corpus for research and development.
\newblock In {\em Proceedings of the 50th Annual Meeting of the Association for
  Computational Linguistics: Short Papers-Volume 2}, pages 203--207.
  Association for Computational Linguistics.

\bibitem[Bordes and Weston, 2016]{bordes2016learning}
Bordes, A. and Weston, J. (2016).
\newblock Learning end-to-end goal-oriented dialog.
\newblock {\em arXiv preprint arXiv:1605.07683}.

\bibitem[Cuay{\'a}huitl, 2017]{cuayahuitl2017simpleds}
Cuay{\'a}huitl, H. (2017).
\newblock Simpleds: A simple deep reinforcement learning dialogue system.
\newblock In {\em Dialogues with Social Robots}, pages 109--118. Springer.

\bibitem[Dhingra et~al., 2016]{dhingra2016end}
Dhingra, B., Li, L., Li, X., Gao, J., Chen, Y.-N., Ahmed, F., and Deng, L.
  (2016).
\newblock End-to-end reinforcement learning of dialogue agents for information
  access.
\newblock {\em arXiv preprint arXiv:1609.00777}.

\bibitem[Dimovski et~al., 2018]{dimovski2018submodularity}
Dimovski, M., Ilievski, V., Musat, C., Hossmann, A., and Baeriswyl, M. (2018).
\newblock Submodularity-inspired data selection for goal-oriented chatbot
  training based on sentence embeddings.
\newblock {\em arXiv preprint arXiv:1802.00757}.

\bibitem[Epstein, 1992]{epstein1992quest}
Epstein, R. (1992).
\newblock The quest for the thinking computer.
\newblock {\em AI magazine}, 13(2):81.

\bibitem[Ga{\v{s}}i{\'c} et~al., 2015]{gavsic2015distributed}
Ga{\v{s}}i{\'c}, M., Kim, D., Tsiakoulis, P., and Young, S. (2015).
\newblock Distributed dialogue policies for multi-domain statistical dialogue
  management.
\newblock In {\em Acoustics, Speech and Signal Processing (ICASSP), 2015 IEEE
  International Conference on}, pages 5371--5375. IEEE.

\bibitem[Hakkani-T{\"u}r et~al., 2016]{hakkani2016multi}
Hakkani-T{\"u}r, D., T{\"u}r, G., Celikyilmaz, A., Chen, Y.-N., Gao, J., Deng,
  L., and Wang, Y.-Y. (2016).
\newblock Multi-domain joint semantic frame parsing using bi directional
  rnn-lstm.
\newblock In {\em INTERSPEECH}, pages 715--719.

\bibitem[Henderson, 2015a]{henderson2015machine}
Henderson, M. (2015a).
\newblock Machine learning for dialog state tracking: A review.
\newblock In {\em Proc. of The First International Workshop on Machine Learning
  in Spoken Language Processing}.

\bibitem[Henderson, 2015b]{henderson:ml-for-dst-review}
Henderson, M. (2015b).
\newblock Machine learning for dialog state tracking: A review.
\newblock In {\em Proceedings of The First International Workshop on Machine
  Learning in Spoken Language Processing}.

\bibitem[Henderson et~al., 2013]{henderson2013dialog}
Henderson, M., Thomson, B., and Williams, J. (2013).
\newblock Dialog state tracking challenge 2 \& 3.

\bibitem[Henderson et~al., 2014a]{henderson2014word}
Henderson, M., Thomson, B., and Young, S. (2014a).
\newblock Word-based dialog state tracking with recurrent neural networks.
\newblock In {\em Proceedings of the 15th Annual Meeting of the Special
  Interest Group on Discourse and Dialogue (SIGDIAL)}, pages 292--299.

\bibitem[Henderson et~al., 2014b]{Henderson2014d}
Henderson, M., Thomson, B., and Young, S.~J. (2014b).
\newblock {Robust Dialog State Tracking Using Delexicalised Recurrent Neural
  Networks and Unsupervised Adaptation}.
\newblock In {\em Proceedings of IEEE Spoken Language Technology}.

\bibitem[Henderson, 2015c]{henderson2015discriminative}
Henderson, M.~S. (2015c).
\newblock {\em Discriminative methods for statistical spoken dialogue systems}.
\newblock PhD thesis, University of Cambridge.

\bibitem[Ilievski et~al., 2018]{ilievski2018goal}
Ilievski, V., Musat, C., Hossmann, A., and Baeriswyl, M. (2018).
\newblock Goal-oriented chatbot dialog management bootstrapping with transfer
  learning.
\newblock {\em arXiv preprint arXiv:1802.00500}.

\bibitem[Kurzweil, 2010]{kurzweil2010singularity}
Kurzweil, R. (2010).
\newblock {\em The singularity is near}.
\newblock Gerald Duckworth \& Co.

\bibitem[Larsson and Traum, 2000]{larsson2000information}
Larsson, S. and Traum, D.~R. (2000).
\newblock Information state and dialogue management in the trindi dialogue move
  engine toolkit.
\newblock {\em Natural language engineering}, 6(3-4):323--340.

\bibitem[Lee, 2017]{lee2017toward}
Lee, S. (2017).
\newblock Toward continual learning for conversational agents.
\newblock {\em arXiv preprint arXiv:1712.09943}.

\bibitem[Li et~al., 2017]{li2017end}
Li, X., Chen, Y.-N., Li, L., and Gao, J. (2017).
\newblock End-to-end task-completion neural dialogue systems.
\newblock {\em arXiv preprint arXiv:1703.01008}.

\bibitem[Li et~al., 2016]{li2016user}
Li, X., Lipton, Z.~C., Dhingra, B., Li, L., Gao, J., and Chen, Y.-N. (2016).
\newblock A user simulator for task-completion dialogues.
\newblock {\em arXiv preprint arXiv:1612.05688}.

\bibitem[Liu and Lane, 2016]{liu2016attention}
Liu, B. and Lane, I. (2016).
\newblock Attention-based recurrent neural network models for joint intent
  detection and slot filling.
\newblock {\em arXiv preprint arXiv:1609.01454}.

\bibitem[Mnih et~al., 2015]{mnih2015human}
Mnih, V., Kavukcuoglu, K., Silver, D., Rusu, A.~A., Veness, J., Bellemare,
  M.~G., Graves, A., Riedmiller, M., Fidjeland, A.~K., Ostrovski, G., et~al.
  (2015).
\newblock Human-level control through deep reinforcement learning.
\newblock {\em Nature}, 518(7540):529--533.

\bibitem[Pan and Yang, 2010]{pan2010survey}
Pan, S.~J. and Yang, Q. (2010).
\newblock A survey on transfer learning.
\newblock {\em IEEE Transactions on knowledge and data engineering},
  22(10):1345--1359.

\bibitem[Peng et~al., 2017]{peng2017composite}
Peng, B., Li, X., Li, L., Gao, J., Celikyilmaz, A., Lee, S., and Wong, K.-F.
  (2017).
\newblock Composite task-completion dialogue policy learning via hierarchical
  deep reinforcement learning.
\newblock In {\em Proceedings of the 2017 Conference on Empirical Methods in
  Natural Language Processing}, pages 2221--2230.

\bibitem[Pieraccini and Huerta, 2005]{pieraccini2005we}
Pieraccini, R. and Huerta, J. (2005).
\newblock Where do we go from here? research and commercial spoken dialog
  systems.
\newblock In {\em 6th SIGdial Workshop on Discourse and Dialogue}.

\bibitem[Schatzmann and Young, 2009]{schatzmann2009hidden}
Schatzmann, J. and Young, S. (2009).
\newblock The hidden agenda user simulation model.
\newblock {\em IEEE transactions on audio, speech, and language processing},
  17(4):733--747.

\bibitem[Serban et~al., 2016]{serban2016building}
Serban, I.~V., Sordoni, A., Bengio, Y., Courville, A.~C., and Pineau, J.
  (2016).
\newblock Building end-to-end dialogue systems using generative hierarchical
  neural network models.
\newblock In {\em AAAI}, pages 3776--3784.

\bibitem[Strayer et~al., 2017]{strayer2017smartphone}
Strayer, D.~L., Cooper, J.~M., Turrill, J., Coleman, J.~R., and Hopman, R.~J.
  (2017).
\newblock The smartphone and the driver’s cognitive workload: A comparison of
  apple, google, and microsoft’s intelligent personal assistants.
\newblock {\em Canadian Journal of Experimental Psychology/Revue canadienne de
  psychologie exp{\'e}rimentale}, 71(2):93.

\bibitem[Su et~al., 2016]{su2016continuously}
Su, P.-H., Gasic, M., Mrksic, N., Rojas-Barahona, L., Ultes, S., Vandyke, D.,
  Wen, T.-H., and Young, S. (2016).
\newblock Continuously learning neural dialogue management.
\newblock {\em arXiv preprint arXiv:1606.02689}.

\bibitem[Sukhbaatar et~al., 2015]{sukhbaatar2015end}
Sukhbaatar, S., Weston, J., Fergus, R., et~al. (2015).
\newblock End-to-end memory networks.
\newblock In {\em Advances in neural information processing systems}, pages
  2440--2448.

\bibitem[Sutskever et~al., 2014]{sutskever2014sequence}
Sutskever, I., Vinyals, O., and Le, Q.~V. (2014).
\newblock Sequence to sequence learning with neural networks.
\newblock In {\em Advances in neural information processing systems}, pages
  3104--3112.

\bibitem[Sutton and Barto, 1998]{sutton1998reinforcement}
Sutton, R.~S. and Barto, A.~G. (1998).
\newblock {\em Reinforcement learning: An introduction}, volume~1.
\newblock MIT press Cambridge.

\bibitem[Van~Hasselt et~al., 2016]{van2016deep}
Van~Hasselt, H., Guez, A., and Silver, D. (2016).
\newblock Deep reinforcement learning with double q-learning.
\newblock In {\em AAAI}, volume~16, pages 2094--2100.

\bibitem[Vinyals and Le, 2015]{vinyals2015neural}
Vinyals, O. and Le, Q. (2015).
\newblock A neural conversational model.
\newblock {\em arXiv preprint arXiv:1506.05869}.

\bibitem[Wang et~al., 2015]{wang2015learning}
Wang, Z., Wen, T.-H., Su, P.-H., and Stylianou, Y. (2015).
\newblock Learning domain-independent dialogue policies via ontology
  parameterisation.

\bibitem[Watkins and Dayan, 1992]{watkins1992q}
Watkins, C.~J. and Dayan, P. (1992).
\newblock Q-learning.
\newblock {\em Machine learning}, 8(3-4):279--292.

\bibitem[Wen et~al., 2016a]{wen2016conditional}
Wen, T.-H., Gasic, M., Mrksic, N., Rojas-Barahona, L.~M., Su, P.-H., Ultes, S.,
  Vandyke, D., and Young, S. (2016a).
\newblock Conditional generation and snapshot learning in neural dialogue
  systems.
\newblock {\em arXiv preprint arXiv:1606.03352}.

\bibitem[Wen et~al., 2015]{wen2015semantically}
Wen, T.-H., Gasic, M., Mrksic, N., Su, P.-H., Vandyke, D., and Young, S.
  (2015).
\newblock Semantically conditioned lstm-based natural language generation for
  spoken dialogue systems.
\newblock {\em arXiv preprint arXiv:1508.01745}.

\bibitem[Wen et~al., 2016b]{wen2016network}
Wen, T.-H., Vandyke, D., Mrksic, N., Gasic, M., Rojas-Barahona, L.~M., Su,
  P.-H., Ultes, S., and Young, S. (2016b).
\newblock A network-based end-to-end trainable task-oriented dialogue system.
\newblock {\em arXiv preprint arXiv:1604.04562}.

\bibitem[Williams et~al.,
  2016]{the-dialog-state-tracking-challenge-series-a-review}
Williams, J., Raux, A., and Henderson, M. (2016).
\newblock The dialog state tracking challenge series: A review.
\newblock {\em Dialogue \& Discourse}.

\bibitem[Williams et~al., 2017]{williams2017hybrid}
Williams, J.~D., Asadi, K., and Zweig, G. (2017).
\newblock Hybrid code networks: practical and efficient end-to-end dialog
  control with supervised and reinforcement learning.
\newblock {\em arXiv preprint arXiv:1702.03274}.

\bibitem[Williams, 1992]{williams1992simple}
Williams, R.~J. (1992).
\newblock Simple statistical gradient-following algorithms for connectionist
  reinforcement learning.
\newblock In {\em Reinforcement Learning}, pages 5--32. Springer.

\bibitem[Yao et~al., 2014]{yao2014spoken}
Yao, K., Peng, B., Zhang, Y., Yu, D., Zweig, G., and Shi, Y. (2014).
\newblock Spoken language understanding using long short-term memory neural
  networks.
\newblock In {\em Spoken Language Technology Workshop (SLT), 2014 IEEE}, pages
  189--194. IEEE.

\bibitem[Young et~al., 2013]{young2013pomdp}
Young, S., Ga{\v{s}}i{\'c}, M., Thomson, B., and Williams, J.~D. (2013).
\newblock Pomdp-based statistical spoken dialog systems: A review.
\newblock {\em Proceedings of the IEEE}, 101(5):1160--1179.

\bibitem[Zen et~al., 2009]{zen2009statistical}
Zen, H., Tokuda, K., and Black, A.~W. (2009).
\newblock Statistical parametric speech synthesis.
\newblock {\em Speech Communication}, 51(11):1039--1064.

\bibitem[Zhang et~al., 2017]{zhang2017towards}
Zhang, Y., Pezeshki, M., Brakel, P., Zhang, S., Bengio, C. L.~Y., and
  Courville, A. (2017).
\newblock Towards end-to-end speech recognition with deep convolutional neural
  networks.
\newblock {\em arXiv preprint arXiv:1701.02720}.

\bibitem[Zhao and Eskenazi, 2016]{zhao2016towards}
Zhao, T. and Eskenazi, M. (2016).
\newblock Towards end-to-end learning for dialog state tracking and management
  using deep reinforcement learning.
\newblock {\em arXiv preprint arXiv:1606.02560}.

\bibitem[Zue et~al., 2000]{zue2000juplter}
Zue, V., Seneff, S., Glass, J.~R., Polifroni, J., Pao, C., Hazen, T.~J., and
  Hetherington, L. (2000).
\newblock Juplter: a telephone-based conversational interface for weather
  information.
\newblock {\em IEEE Transactions on speech and audio processing}, 8(1):85--96.

\end{thebibliography}
\addcontentsline{toc}{chapter}{Bibliography}

\end{document}